\documentclass{article}

 \usepackage[preprint]{neurips_2026}

\usepackage[utf8]{inputenc} 
\usepackage[T1]{fontenc}    
\usepackage{hyperref}       
\hypersetup{hidelinks,breaklinks=true}
\usepackage{url}            
\usepackage{booktabs}       
\usepackage{amsfonts}       
\usepackage{amsmath}
\usepackage{nicefrac}       
\usepackage{microtype}      
\usepackage{xcolor}         
\usepackage{multirow}
\usepackage{multicol}
\usepackage{makecell}
\usepackage{graphicx}
\usepackage{subcaption}
\usepackage{wrapfig}
\usepackage{float}
\usepackage{placeins}
\usepackage{booktabs}
\usepackage{multirow}
\usepackage{array}
\usepackage{ragged2e}
\usepackage{fvextra}
\newcolumntype{L}[1]{>{\RaggedRight\arraybackslash}p{#1}}
\DefineVerbatimEnvironment{promptbox}{Verbatim}{
  fontsize=\footnotesize,
  breaklines=true,
  breaksymbolleft={},
  breakindent=1em,
  commandchars=\\\{\}
}

\title{BehaviorBench: Modeling Real-World User Decisions from Behavioral Traces}

\author{%
Liangwei Yang, Jielin Qiu, Zixiang Chen, Ming Zhu, Juntao Tan, Zhiwei Liu \\
\textbf{Wenting Zhao, Zhujun Lan, Akshara Prabhakar, Silvio Savarese, Huan Wang, Shelby Heinecke} \\
Salesforce AI Research
}

\begin{document}

\maketitle

\begin{abstract}
Many decision-support settings require systems that adapt to individual users, but evaluation data for this problem remain limited. Existing benchmarks for user understanding often rely on simulated users or model-generated behavior, even though recent work cautions that model-based simulations can diverge systematically from human behavior. We introduce \textsc{BehaviorBench}, a benchmark for evaluating personalized decision modeling from real-world behavioral traces. \textsc{BehaviorBench} reconstructs wallet-level decision histories from observed public prediction-market and on-chain records, and organizes them into two complementary task layers: \emph{Belief prediction}, which predicts a user's final revealed stance and confidence in a market, and \emph{Trade prediction}, which predicts the direction and amount of individual transactions. Across 2,000 evaluation wallets, the benchmark contains 141,445 Belief instances and 1,485,972 Trade instances, with disjoint support pools for retrieval-based evaluation. We evaluate frontier and open-weight generative models under four history interfaces: no personalization, direct recent history, generated user profiles, and retrieved support-wallet evidence. Personalization improves Belief prediction more consistently than Trade prediction, model rankings change across task layers and metrics, and different history interfaces expose different failure modes. \textsc{BehaviorBench} provides an evaluation setting for studying whether personalized methods can use real-world behavioral evidence rather than simulated users alone.
\end{abstract}

\section{Introduction}

Tool-using models and agents are now commonly evaluated as general-purpose tools: they navigate websites, call APIs, edit repositories, or repair software \citep{zhouwebarena,yao2025tau,jimenez2023swe,zhang2024autocoderover}. This tool-centric view has produced useful progress, but it also treats the user mainly as a source of instructions. As agents become more deeply embedded in everyday decision support, the next question is not only whether they can complete a task, but whether they can understand the person for whom the task is being done. A useful personalized agent should infer how a particular person tends to believe, choose, and act from that person's prior behavior, rather than applying the same generic decision rule to everyone. This motivation connects to recent work on personalization and preference following \citep{salemi2024lamp,zhao2025llms}, dynamic user profiling \citep{jiang2025know,wu2026knowme}, and long-term agent memory \citep{packer2023memgpt,zhong2024memorybank,chhikara2025mem0}.

This capability is hard to evaluate because the relevant user signal is rarely stated directly. Real preferences, beliefs, and decision tendencies are hidden in sequences of behavior: what a person repeatedly chooses, avoids, revises, accumulates, or abandons. Existing personalization benchmarks often study stated personas, dialogue histories, annotated profiles, or constructed preferences \citep{zhang2018personalizing,zheng2019personalized,tan2025personabench,tao2025personafeedback}, while memory benchmarks test whether systems can retrieve or maintain user-specific information over time \citep{zhang2026memorycd,hu2025memory}. These settings are valuable, but a system may complete an assigned task correctly while still failing to infer what a specific user will believe, prefer, or do next from implicit behavior. Measuring this failure requires benchmarks grounded in real-world behavioral traces rather than only in explicit personas or one-off survey answers.

Existing evaluation data for this problem remain limited. A growing line of work studies simulated users and generative social agents \citep{park2023generative,li2025behaviorchain,zhang2024usimagent}, including agent-based simulators for recommendation and other behavioral domains \citep{zhang2024agent4rec,wang2024recagent}. But simulation is not a substitute for behavioral evidence. Recent benchmarks and analyses caution that plausible synthetic behavior need not match real human behavior \citep{hu2025simbench,lee2025realtalk,zhou2026sim2real,zhu2026realusersim}. These findings motivate benchmarks grounded in observed behavior rather than only in model-generated personas.

Public event-market traces provide a useful real-world setting for this purpose. They record repeated decisions made by pseudonymous users under changing evidence and incentives. Unlike static survey answers, these traces contain temporally ordered actions tied to concrete event contexts. Unlike private application logs, they can be reconstructed from public data sources and documented for reproducible benchmark construction.

We introduce \textsc{BehaviorBench}, a benchmark for modeling user decisions from real-world behavioral traces. The benchmark tests how different decision systems and history interfaces use prior behavior to predict user decisions at two levels of abstraction. The \emph{Belief prediction} layer asks whether a system can infer a user's final revealed stance and confidence in an event. The \emph{Trade prediction} layer asks whether a system can predict the direction and amount of an individual next action. This separation is important: final beliefs reflect relatively stable revealed positions, while local actions are shaped by timing, exposure, and changing context.

The benchmark also treats personalization as an evaluation design choice. We study three generation interfaces that expose different forms of behavioral evidence: DirectGen uses recent user history, ProfileGen uses a generated structured user profile, and RetrievalGen uses retrieved behavioral evidence from a disjoint support pool. These interfaces support different personalization regimes, from short-horizon sequential prediction to compact user summarization and cross-user behavioral analogy, while keeping the underlying decision system separate from the representation of user history.

The resulting empirical picture is that personalization is not a single capability. Different behavioral targets reward different representations of history: profile summaries are useful for final revealed beliefs, while transaction-level trades depend more strongly on concrete sequential context.

Our main contributions are:
\begin{itemize}
    \item We introduce \textsc{BehaviorBench}, a benchmark for personalized decision modeling from real-world public behavioral traces, with 141,445 Belief instances and 1,485,972 Trade instances across 2,000 evaluation entities.
    \item We reconstruct high-fidelity longitudinal decision histories from observed on-chain behavior and document a quality-controlled cohort construction pipeline for producing stable benchmark users from raw logs.
    \item We define two complementary task layers, Belief prediction and Trade prediction, and four evaluation interfaces that test whether models benefit from no history, direct history, generated profiles, or retrieved behavioral evidence.
    \item We benchmark frontier and open-weight generative models, showing that personalization is useful but strategy-dependent, and that rankings vary substantially across task layer and metric.
\end{itemize}

\section{BehaviorBench Tasks}

\paragraph{Problem definition.}

Let $\mathcal{U}$ be the set of pseudonymous wallets and $\mathcal{M}$ the set of markets. Each market $m \in \mathcal{M}$ has an outcome set $\mathcal{O}_m$; in the binary case these correspond to YES and NO positions. We represent a reconstructed trade event as
\[
e_i=(u_i,m_i,o_i,d_i,a_i,b_i,\tau_i),
\]
where $u_i \in \mathcal{U}$ is the wallet, $m_i$ is the market, $o_i \in \mathcal{O}_{m_i}$ is the outcome token, $d_i \in \{+1,-1\}$ is the signed direction with $+1$ for BUY and $-1$ for SELL, $a_i \geq 0$ is the parsed token amount, $b_i$ is the block number, and $\tau_i$ is the within-block transaction/log order used to sort events. For a target index $i$, $h_{u_i,<i}$ denotes the visible behavior of wallet $u_i$ before the target event or target market decision.

Each benchmark instance contains a target context $x_i$ describing the relevant event, market, or transaction context and, depending on the evaluation interface, a representation of prior behavior. A decision system produces a prediction $\hat{y}=f(x_i, h_{u_i,<i})$ for a held-out user decision $y_i$.

\textsc{BehaviorBench} evaluates two levels of behavioral abstraction. The Belief prediction layer treats a wallet--market pair as the unit of prediction and asks for the user's final revealed position in that market. The Trade prediction layer treats an individual transaction as the unit of prediction and asks for the user's local trading action. Evaluating both layers avoids collapsing stable revealed beliefs and short-horizon actions into a single target.

\paragraph{Belief prediction.}

For each wallet--market pair $(u,m)$, the Belief layer aggregates the wallet's transaction sequence into final outcome-level positions and asks for the market-level final stance. Let
\[
P_{u,m,o}=\sum_{i:u_i=u,\,m_i=m,\,o_i=o} d_i a_i
\]
be the reconstructed final net position of wallet $u$ in outcome $o$ of market $m$ after applying all reconstructed trades in the wallet--market trajectory. We define the final revealed side as the outcome with the largest nonnegative final position,
\[
y^{\mathrm{side}}_{u,m}=\arg\max_{o\in\mathcal{O}_m} \max(P_{u,m,o},0).
\]
For retained benchmark instances, this final side is well-defined after preprocessing; any degenerate ties are resolved deterministically by the released label-construction script. The Belief confidence label measures how concentrated the final position is on the dominant outcome. Let $Z_{u,m}=\sum_{o\in\mathcal{O}_m}\max(P_{u,m,o},0)$. We define
\[
y^{\mathrm{conf}}_{u,m}=
\begin{cases}
\frac{\max_{o\in\mathcal{O}_m}\max(P_{u,m,o},0)}{Z_{u,m}}, & Z_{u,m}>0,\\
0, & Z_{u,m}=0,
\end{cases}
\]
so confidence is the share of the final nonnegative position held in the dominant outcome. In the released benchmark, confidence is reported on the $[0,1]$ scale. This target is a revealed behavioral signal, not a direct measurement of a private psychological belief.

\paragraph{Trade prediction.}

For each transaction-level target event $e_i$, the Trade layer predicts the direction and amount of the next local action:
\[
y^{\mathrm{dir}}_i =
\begin{cases}
\mathrm{BUY}, & d_i=+1,\\
\mathrm{SELL}, & d_i=-1,
\end{cases}
\qquad
y^{\mathrm{amt}}_i=a_i.
\]
Amount $a_i$ is parsed from the raw on-chain amount field into an integer-valued trade size. We also derive an auxiliary action label from the signed position change $\Delta_i=d_i a_i$. Let $Q_{u_i,m_i,o_i}^{<i}$ denote the position in the same wallet--market--outcome sequence immediately before event $i$, and let $Q_{u_i,m_i,o_i}^{i}=Q_{u_i,m_i,o_i}^{<i}+\Delta_i$ be the position after the event. The action label is
\[
y^{\mathrm{act}}_i =
\begin{cases}
\mathrm{accumulate}, & \Delta_i > 0,\\
\mathrm{reduce}, & \Delta_i < 0 \ \text{and}\ Q_{u_i,m_i,o_i}^{i}>0,\\
\mathrm{close}, & \Delta_i < 0 \ \text{and}\ Q_{u_i,m_i,o_i}^{i}\leq 0.
\end{cases}
\]
The main Trade evaluation reports direction and amount; the action label is used to interpret the type of position change.

\paragraph{Evaluation metrics.}

Belief prediction is evaluated with final-side Choice Accuracy and Confidence MAE on the $[0,1]$ scale. Trade prediction is evaluated with BUY/SELL Direction Accuracy and amount Median Absolute Error. Accuracy metrics capture directional behavioral alignment, while error metrics capture calibration and magnitude sensitivity. Appendix~\ref{app:metric_formulas} gives the exact formulas used in the released evaluation scripts.

\paragraph{History interfaces and model instantiation.}

We evaluate each task under history interfaces that expose different forms of behavioral evidence. The no-personalization baseline uses only the target context. DirectGen uses recent visible history, ProfileGen uses a structured profile generated from pre-target behavior, and RetrievalGen supplies retrieved evidence from a disjoint support pool. We run the same benchmark interface across frontier proprietary and open-weight generative models; other decision systems or agent architectures can consume the same input fields and output schemas. Appendix~\ref{app:prompt_schema} gives the exact input schemas, history construction details, decoding settings, and parse-failure handling.

\section{Dataset Construction}

\paragraph{Data sources.}

\begin{figure*}[t]
\centering
\includegraphics[width=\textwidth]{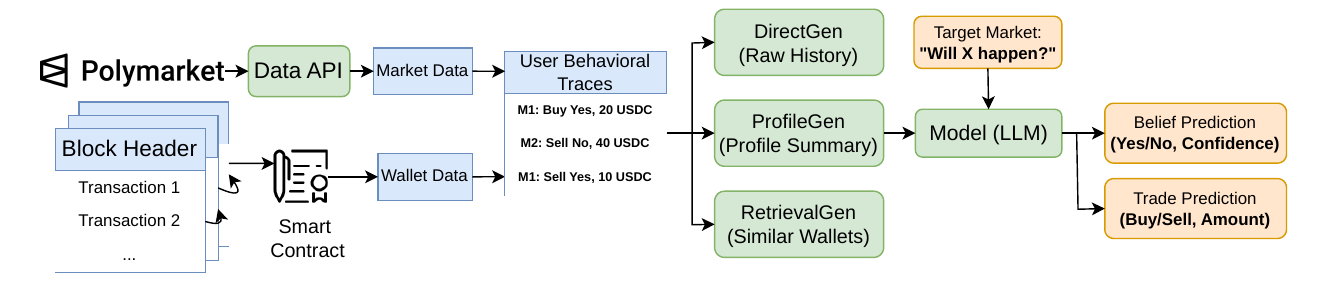}
\caption{Overview of the \textsc{BehaviorBench} construction pipeline. We collect public event metadata and on-chain logs, reconstruct wallet--market--outcome behavior, apply cohort and quality-control filters, and build two benchmark layers for Belief prediction and Trade prediction with disjoint support pools for retrieval-based evaluation.}
\label{fig:benchmark_pipeline}
\end{figure*}

\textsc{BehaviorBench} is constructed from real-world public event-market behavioral traces and public on-chain logs. Market and event metadata are collected from public APIs, while wallet--market--outcome trading trajectories are reconstructed from Polygon on-chain logs. In this setting, a wallet is a pseudonymous trading entity and serves as the behavioral unit for user modeling; it is not treated as a verified real-world identity. For provenance and reproducibility, the metadata source is the Polymarket public API; the rest of the paper treats the artifact as a benchmark of real-world public behavioral traces rather than as a study of any specific platform.
Figure~\ref{fig:benchmark_pipeline} summarizes the end-to-end construction pipeline from public records to benchmark-ready Belief and Trade prediction instances.

\paragraph{Raw data collection.}

The raw on-chain pipeline expands wallet identifiers from \texttt{topic\_addresses} and computes wallet-level activity features from observed logs, including log count, distinct transaction count, distinct contract count, and active-block count. Trade reconstruction parses the raw direction and amount fields, converts \texttt{amount\_hex} into integer amounts, and orders each wallet--market--outcome sequence by block and transaction position to recover position changes over time.

\paragraph{Cohort construction and quality control.}

We form a repeatedly active wallet cohort from public on-chain logs, beginning with candidate entities expanded from \texttt{topic\_addresses} and filtered by basic activity features such as log count, distinct transactions, distinct contracts, and active blocks. We then require at least 10 interacted markets, yielding a common starting cohort of 31,319 wallets. The final benchmark applies quality-control filters that remove inconsistent negative-position trajectories, sparse user--market histories, highly concentrated or bursty activity, and extreme sequence lengths. We cap each user--market sequence at 30 rows.

The main benchmark is restricted to block $\leq 80{,}000{,}000$, corresponding to a data vintage ending in December 2025. We use this cutoff to avoid a visible regime shift in the raw collection: transaction volume increases sharply beginning in January 2026, plausibly reflecting more automated or agent-mediated activity. The cutoff is a benchmark-design choice for temporal comparability, not a claim that later activity is invalid. Appendix~\ref{app:cohort_filtering} gives the complete threshold chain.

\paragraph{Behavior reconstruction and splits.}

After cohort filtering, we transform reconstructed behavior into two aligned benchmark layers. The Belief prediction layer aggregates each wallet--market trajectory into a final side and confidence target. The Trade prediction layer preserves individual transaction-level actions and constructs instances around the target trade. Both layers are split chronologically into train, validation, and test partitions so that target labels are never available when constructing the model input. Retrieval-based evaluation additionally uses a disjoint support pool of wallets, preventing target-wallet identity leakage through retrieved examples.

\paragraph{Dataset statistics.}

Table~\ref{tab:bucket_split_tx_stats_with_layer_specific_support} summarizes the benchmark scale. We sample 2,000 evaluation wallets using task-specific activity buckets so the benchmark covers heterogeneous users rather than only the most common activity pattern. Belief buckets are based on markets per wallet, while Trade buckets are based on average transactions per market. The resulting benchmark contains 141,445 Belief instances and 1,485,972 Trade instances, plus disjoint support pools for retrieval-based evaluation.

\begin{table*}[t]
\centering
\footnotesize
\setlength{\tabcolsep}{5pt}
\renewcommand{\arraystretch}{1.05}
\caption{Bucket-wise wallet and instance (Inst.) statistics for Belief prediction and Trade prediction, with layer-specific support sets for retrieval augmented generation.}
\label{tab:bucket_split_tx_stats_with_layer_specific_support}
\begin{tabular}{@{}l r r r r r@{}}
\toprule
\textbf{Bucket definition} & \textbf{\#Wallets} & \textbf{\#Train} & \textbf{\#Val} & \textbf{\#Test} & \textbf{\#Total} \\
\midrule
\multicolumn{6}{@{}l}{\textit{Belief prediction: buckets by markets per wallet}} \\
21--49 markets & 1,000 & 21,666 & 4,694 & 5,261 & 31,621 \\
50--99 markets & 600 & 28,102 & 6,067 & 6,356 & 40,525 \\
100--199 markets & 300 & 28,807 & 6,192 & 6,346 & 41,345 \\
200--500 markets & 100 & 19,519 & 4,194 & 4,241 & 27,954 \\
\addlinespace[1pt]
\textbf{Experiment set total} & \textbf{2,000} & \textbf{98,094} & \textbf{21,147} & \textbf{22,204} & \textbf{141,445} \\
Support set & 11,676 & -- & -- & -- & 5,086,400 \\
\midrule
\multicolumn{6}{@{}l}{\textit{Trade prediction: buckets by average transactions per market}} \\
3--5 avg tx/market & 800 & 199,939 & 13,746 & 7,468 & 221,153 \\
5--8 avg tx/market & 800 & 672,319 & 57,257 & 29,711 & 759,287 \\
8--12 avg tx/market & 200 & 267,415 & 25,792 & 13,292 & 306,499 \\
12+ avg tx/market & 200 & 172,268 & 17,676 & 9,089 & 199,033 \\
\addlinespace[1pt]
\textbf{Experiment set total} & \textbf{2,000} & \textbf{1,311,941} & \textbf{114,471} & \textbf{59,560} & \textbf{1,485,972} \\
Support set & 11,676 & -- & -- & -- & 9,986,961 \\
\bottomrule
\end{tabular}
\end{table*}

\paragraph{Artifact documentation and release.}

\textsc{BehaviorBench} is an evaluation artifact for studying how decision systems use behavioral evidence; it is not intended to characterize all prediction-market users or infer private beliefs. The benchmark files are available through Hugging Face\footnote{\url{https://huggingface.co/datasets/Anony84/BehaviorBench}} under the Apache-2.0 license. The release includes benchmark-ready Belief and Trade files, retrieval support pools, preprocessing and evaluation scripts, dataset documentation, and Responsible AI metadata covering provenance, intended use, privacy considerations, and limitations.

\section{Main Results}

\paragraph{Overall performance.}

\begin{table*}[t]
\centering
\footnotesize
\setlength{\tabcolsep}{4.5pt}
\renewcommand{\arraystretch}{1.08}
\caption{Belief prediction results. Best is in bold; second best is underlined.}
\label{tab:belief_results_no_qwen_with_gemma}
\begin{tabular}{@{}l ccc ccc@{}}
\toprule
& \multicolumn{3}{c}{Choice Acc. (\%) $\uparrow$} & \multicolumn{3}{c}{Confidence MAE $\downarrow$} \\
\cmidrule(lr){2-4}\cmidrule(lr){5-7}
Model & DirectGen & ProfileGen & RetrievalGen & DirectGen & ProfileGen & RetrievalGen \\
\midrule
\multicolumn{1}{@{}l}{No Personalization} & \multicolumn{3}{c}{53.13} & \multicolumn{3}{c@{}}{0.3417} \\
\midrule
\multicolumn{7}{@{}l}{\textit{Frontier models}} \\
gpt-5\_4 & \textbf{70.75} & \textbf{73.20} & \textbf{71.03} & 0.3040 & 0.1835 & 0.3098 \\
gpt-5\_4-mini & 66.34 & 69.24 & 66.96 & 0.2827 & 0.1259 & 0.2672 \\
gpt-5\_4-nano & 65.80 & 68.62 & 66.80 & 0.3232 & 0.1515 & 0.3049 \\
claude\_opus\_4 & \underline{70.22} & \underline{71.47} & \underline{70.22} & \underline{0.1189} & \underline{0.0788} & \textbf{0.1131} \\
claude\_3\_7\_sonnet & 67.68 & 71.18 & 69.61 & 0.2540 & \textbf{0.0783} & 0.1964 \\
\midrule
\multicolumn{7}{@{}l}{\textit{Open-weight models}} \\
gemma3\_12b\_it & 70.21 & 71.17 & 70.06 & \textbf{0.1054} & 0.0814 & \underline{0.1256} \\
gemma3\_27b\_it & 70.08 & 70.73 & 69.89 & 0.1864 & 0.0884 & 0.1864 \\
gpt\_oss\_20b & 66.00 & 66.80 & 62.76 & 0.2031 & 0.1020 & 0.2524 \\
llama4\_scout\_17b\_16e\_instruct & 64.77 & 65.88 & 65.12 & 0.1837 & 0.1292 & 0.1986 \\
ministral\_3\_14b\_instruct\_2512 & 59.82 & 62.18 & 60.43 & 0.2568 & 0.1877 & 0.2548 \\
nemotron\_nano\_30b\_a3b\_bf16 & 58.63 & 51.39 & 60.79 & 0.2225 & 0.2037 & 0.1706 \\
\bottomrule
\end{tabular}
\end{table*}

\begin{table*}[t]
\centering
\footnotesize
\setlength{\tabcolsep}{4.5pt}
\renewcommand{\arraystretch}{1.08}
\caption{Trade prediction results. Best is in bold; second best is underlined.}
\label{tab:trade_results_group_by_metric_ranked}
\begin{tabular}{@{}l ccc ccc@{}}
\toprule
& \multicolumn{3}{c}{Direction Acc. (\%) $\uparrow$} & \multicolumn{3}{c}{Median AE $\downarrow$} \\
\cmidrule(lr){2-4}\cmidrule(lr){5-7}
Model & DirectGen & ProfileGen & RetrievalGen & DirectGen & ProfileGen & RetrievalGen \\
\midrule
\multicolumn{1}{@{}l}{No Personalization} & \multicolumn{3}{c}{52.85} & \multicolumn{3}{c@{}}{31.79} \\
\midrule
\multicolumn{7}{@{}l}{\textit{Frontier models}} \\
gpt-5\_4 & \textbf{75.38} & \underline{69.11} & \textbf{68.86} & 9.6317 & 14.6095 & 10.0000 \\
gpt-5\_4-mini & 72.45 & 67.61 & 65.07 & 9.8000 & 12.9600 & 11.1300 \\
gpt-5\_4-nano & 59.79 & 57.94 & 50.38 & 11.3721 & 20.8300 & 10.2233 \\
claude\_opus\_4 & \underline{75.00} & \textbf{69.57} & 63.92 & 12.4744 & 15.3525 & 22.6667 \\
claude\_3\_7\_sonnet & 73.17 & 67.58 & \underline{66.67} & 9.4900 & 14.1152 & \underline{9.7597} \\
\midrule
\multicolumn{7}{@{}l}{\textit{Open-weight models}} \\
gemma3\_12b\_it & 63.63 & 64.57 & 59.82 & \textbf{8.3390} & \underline{10.9850} & \textbf{8.2947} \\
gemma3\_27b\_it & 64.67 & 61.35 & 60.55 & \underline{9.0000} & 15.1833 & 14.6472 \\
gpt\_oss\_20b & 60.57 & 60.69 & 59.99 & 10.4000 & 15.5000 & 14.5000 \\
llama4\_scout\_17b\_16e\_instruct & 70.09 & 63.47 & 60.46 & 9.1447 & \textbf{10.3000} & 10.8134 \\
ministral\_3\_14b\_instruct\_2512 & 48.04 & 53.69 & 50.67 & 21.9900 & 35.4759 & 21.8000 \\
nemotron\_nano\_30b\_a3b\_bf16 & 64.72 & 62.66 & 60.20 & 9.9509 & 17.0740 & 11.5262 \\
\bottomrule
\end{tabular}
\end{table*}

Tables~\ref{tab:belief_results_no_qwen_with_gemma} and~\ref{tab:trade_results_group_by_metric_ranked} report the main benchmark results. Behavioral history provides a clear personalization gain on both task layers: history-aware interfaces consistently outperform the no-personalization baseline. These improvements show that prior behavior contains usable personalized signal, but the task is far from saturated: accuracies remain well below ceiling and continuous error metrics remain substantial.

The two task layers favor different representations of history. ProfileGen is especially effective for Belief, where compressed user summaries improve final-side accuracy and confidence calibration for several models. Trade behaves differently: DirectGen is strongest for direction prediction and amount error for many models, suggesting that transaction-level behavior depends more on concrete sequential context than on abstract user summaries. Thus, personalization in \textsc{BehaviorBench} is not a single capability; the useful form of history depends on the behavioral target.

Model rankings also depend on the metric. The strongest model for side or direction accuracy is not always the strongest model for confidence or amount error. For example, \texttt{gpt-5\_4} is strongest on Belief choice accuracy and Trade direction accuracy, while \texttt{gemma3\_12b\_it} is highly competitive on continuous error metrics, including Trade amount error. A single accuracy leaderboard would therefore miss calibration and magnitude failures that matter for personalized decision modeling.

\paragraph{Model-level comparison.}

\begin{figure*}[t]
\centering
\begin{subfigure}[t]{0.49\textwidth}
\centering
\includegraphics[width=\linewidth]{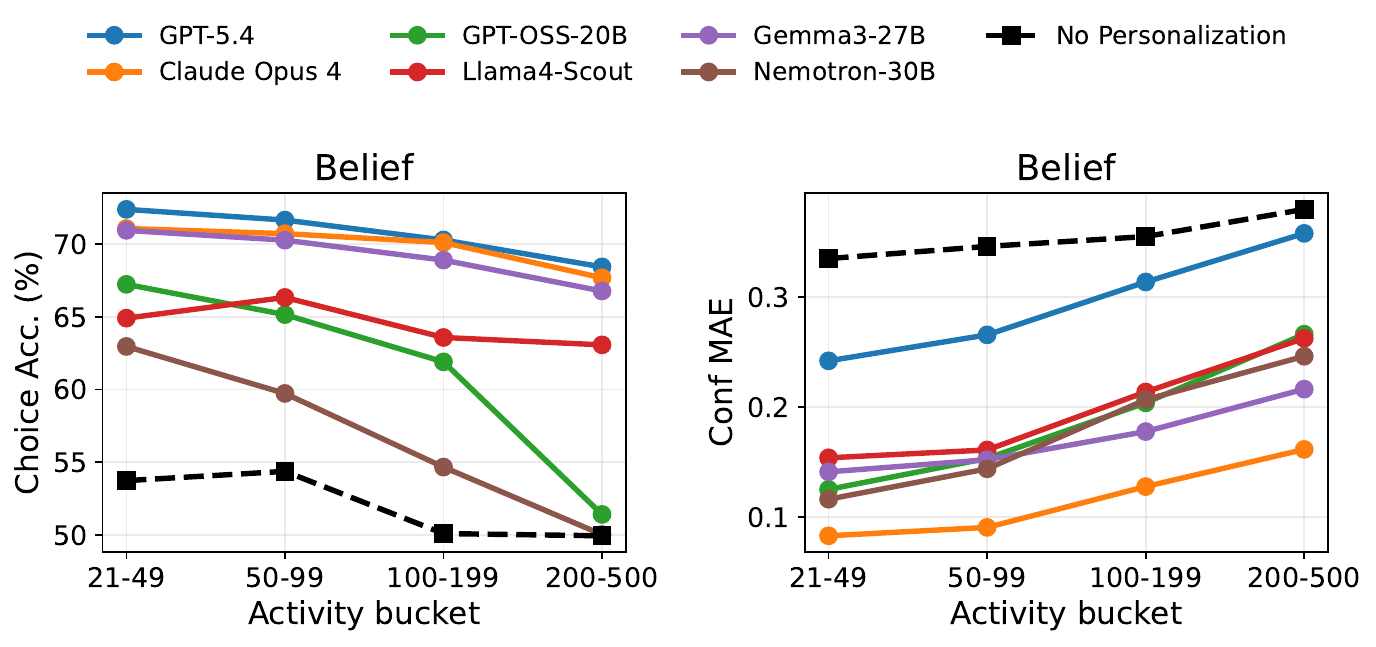}
\caption{Belief prediction.}
\label{fig:main_belief_model}
\end{subfigure}
\hfill
\begin{subfigure}[t]{0.49\textwidth}
\centering
\includegraphics[width=\linewidth]{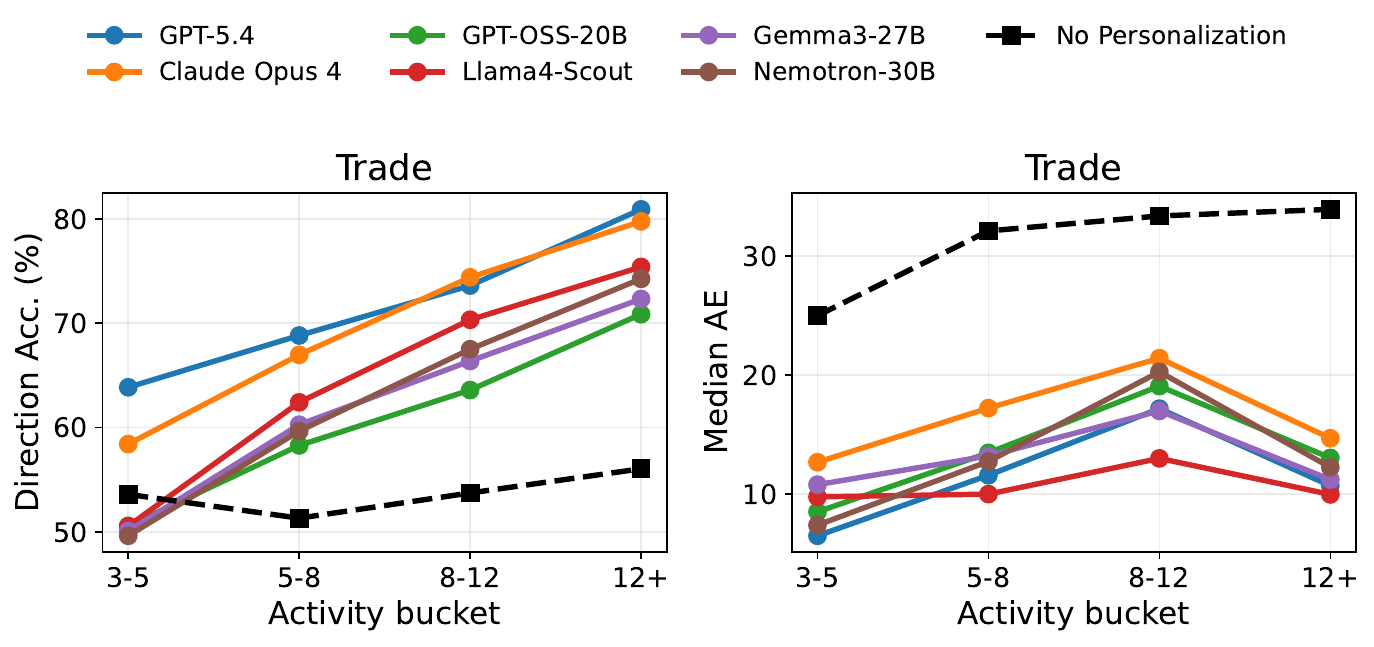}
\caption{Trade prediction.}
\label{fig:main_trade_model}
\end{subfigure}
\caption{Average model performance across history interfaces.}
\label{fig:model_comparison}
\end{figure*}

Figure~\ref{fig:model_comparison} compares model-level performance after averaging over generation interfaces. Frontier models generally lead on choice and direction accuracy, but the ranking is less monotonic on continuous error metrics. Some open-weight models are highly competitive on calibration and amount prediction. This split indicates that model scale or frontier status does not uniformly translate into better behavioral calibration.

\paragraph{History interface comparison.}

\begin{figure*}[t]
\centering
\begin{subfigure}[t]{0.49\textwidth}
\centering
\includegraphics[width=\linewidth]{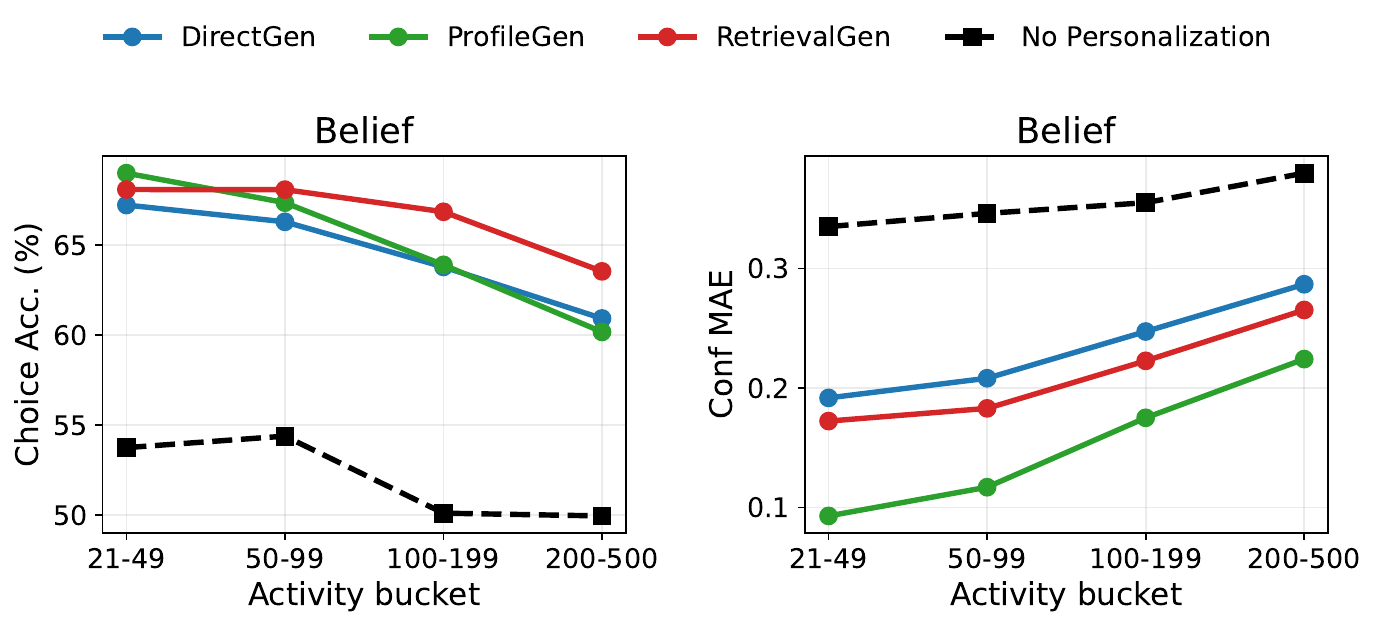}
\caption{Belief prediction.}
\label{fig:main_belief_method}
\end{subfigure}
\hfill
\begin{subfigure}[t]{0.49\textwidth}
\centering
\includegraphics[width=\linewidth]{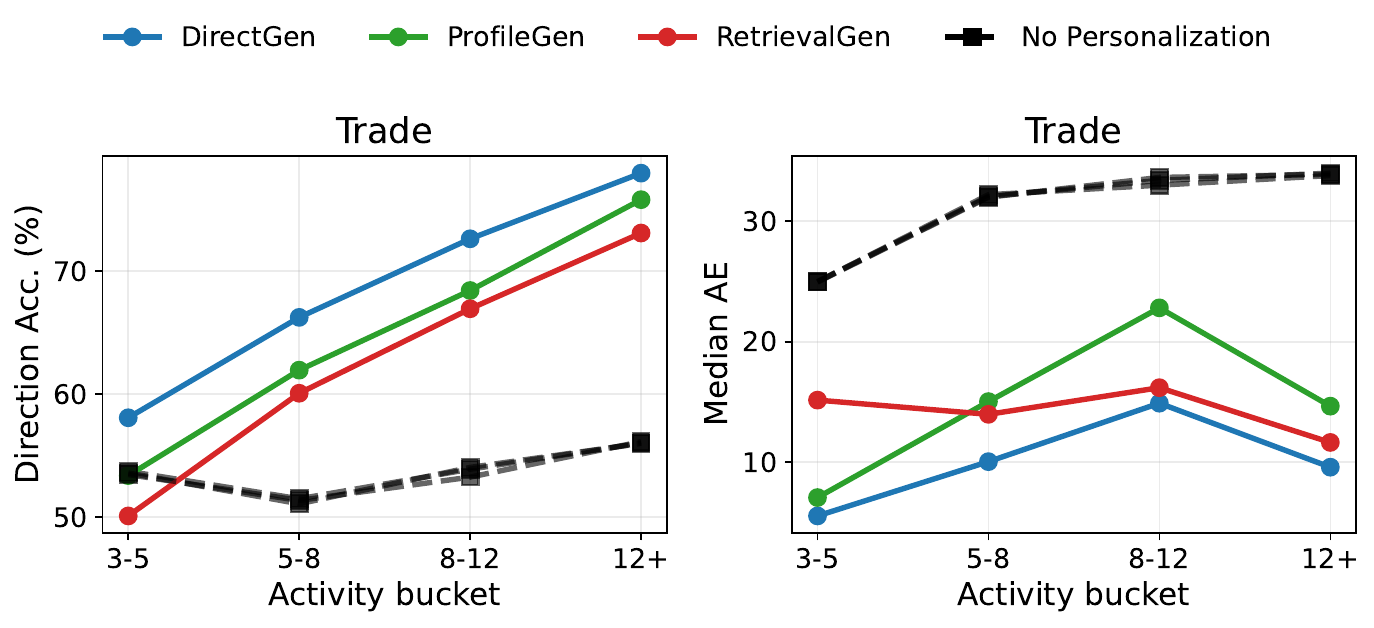}
\caption{Trade prediction.}
\label{fig:main_trade_method}
\end{subfigure}
\caption{Average performance of DirectGen, ProfileGen, and RetrievalGen.}
\label{fig:method_comparison}
\end{figure*}

Figure~\ref{fig:method_comparison} compares the history interfaces. Averaged across models, ProfileGen is strongest for Belief prediction and substantially improves confidence calibration. This suggests that compressed user profiles capture stable tendencies relevant to final revealed positions. Trade shows the opposite structure: DirectGen is strongest on direction and amount prediction, while ProfileGen and RetrievalGen lag behind. Transaction-level behavior therefore appears more dependent on recent local context than on abstract user summaries.

The contrast between Belief and Trade is the central empirical finding of the benchmark. Personalization helps, but not every personalization interface helps every behavioral target. A history representation that is useful for inferring a user's final stance can be too coarse for predicting the next trade.

\section{Analysis}

\paragraph{Belief and Trade measure different capabilities.}

\begin{wrapfigure}{r}{0.48\textwidth}
\vspace{-1.2em}
\centering
\includegraphics[width=0.46\textwidth]{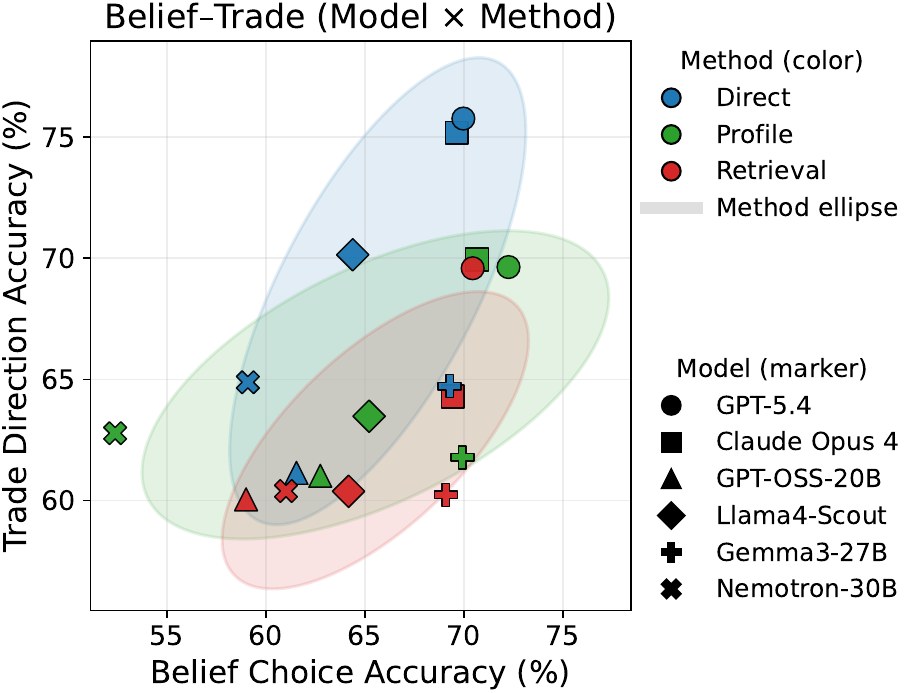}
\caption{Belief--Trade scatter across model--method pairs. Each point is one model $\times$ generation method. Colors denote methods; marker shapes denote models.}
\label{fig:belief_trade_scatter}
\vspace{-1.0em}
\end{wrapfigure}

Figure~\ref{fig:belief_trade_scatter} compares Belief and Trade accuracy across model--method pairs. Strong Belief performance does not automatically imply strong Trade performance: ProfileGen often improves Belief accuracy, while the best Trade results come from DirectGen. This decoupling reflects the difference between predicting a user's final revealed stance and predicting a local action shaped by timing, exposure, and market context.

Bucket-wise decomposed results in Figures~\ref{fig:decomposed_belief} and~\ref{fig:decomposed_trade} show the same pattern from another angle. Broader market coverage does not make Belief prediction easier; accuracy generally declines in higher market-count buckets. Trade direction accuracy, in contrast, improves as average transactions per market increases, suggesting that repeated within-market behavior provides useful local signal. The appendix provides the full decomposed error plots.

\paragraph{How much history helps.}

\begin{figure*}[t]
\centering
\begin{subfigure}[t]{0.4\textwidth}
\centering
\includegraphics[width=\linewidth]{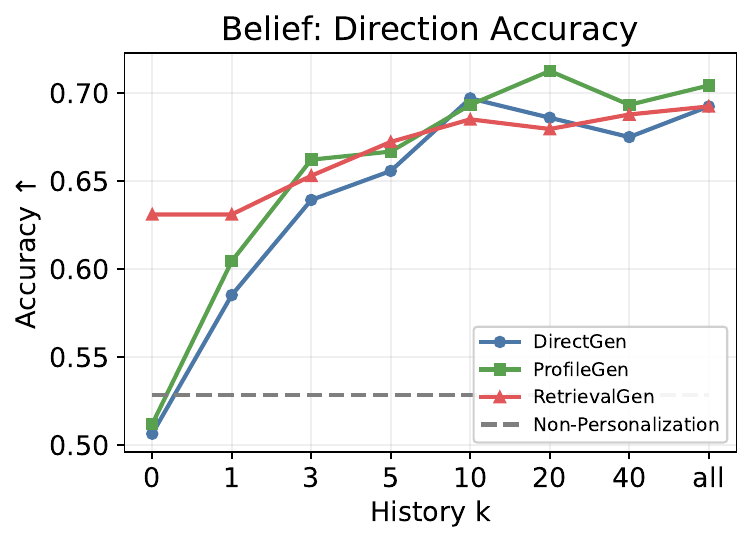}
\end{subfigure}
\begin{subfigure}[t]{0.4\textwidth}
\centering
\includegraphics[width=\linewidth]{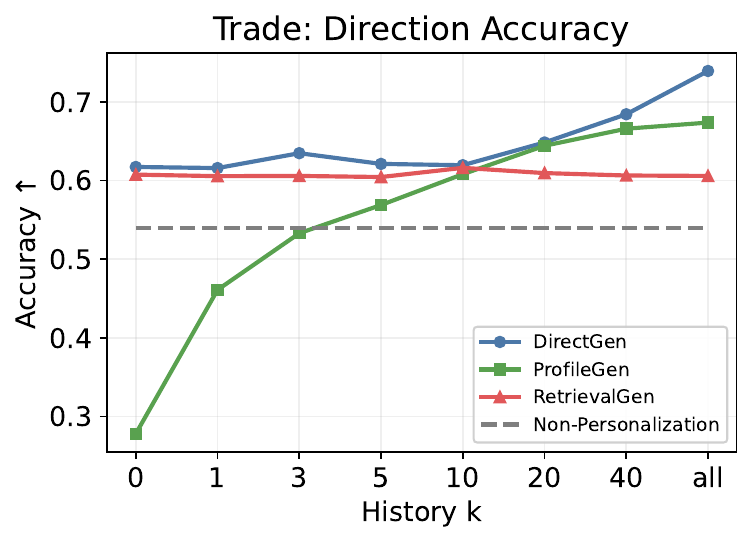}
\end{subfigure}
\caption{History scaling experiments for Belief and Trade direction accuracy.}
\label{fig:history_scaling_main}
\end{figure*}

The history-scaling experiments in Figure~\ref{fig:history_scaling_main} test whether additional behavioral history improves prediction. These experiments are diagnostic rather than the main benchmark ranking. In Belief prediction, accuracy rises sharply when moving from no history to a modest history window, and ProfileGen benefits once enough behavior is available to summarize stable tendencies. Confidence error shows the same pattern: profile summaries help substantially, but very long histories can eventually degrade calibration. This supports the idea that Belief prediction benefits from enough history to summarize stable preferences, but not necessarily from unbounded history.

Trade behaves differently. Direction accuracy for DirectGen increases more steadily with additional history and remains strongest when the full visible history is available. ProfileGen also improves with more history, but does not catch up to direct sequential context, while RetrievalGen is comparatively flat. Amount error mirrors this trend. These results suggest that local transaction prediction benefits from retaining concrete recent or complete histories rather than compressing behavior into a static profile.

\paragraph{Ablations of profile and retrieval.}

\begin{figure*}[t]
\centering
\begin{subfigure}[t]{0.24\textwidth}
\centering
\includegraphics[width=\linewidth]{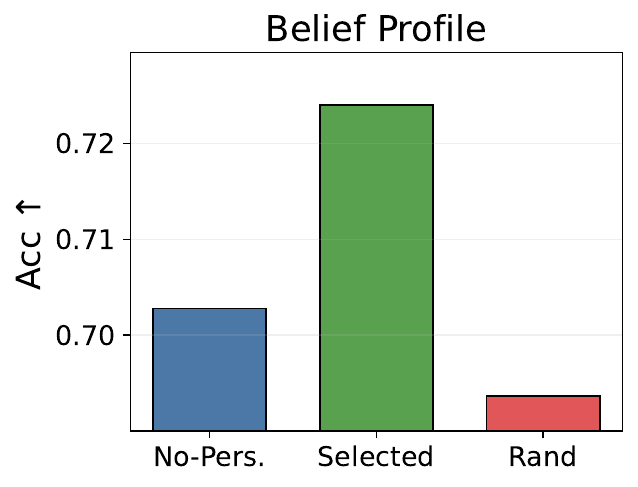}
\end{subfigure}
\begin{subfigure}[t]{0.24\textwidth}
\centering
\includegraphics[width=\linewidth]{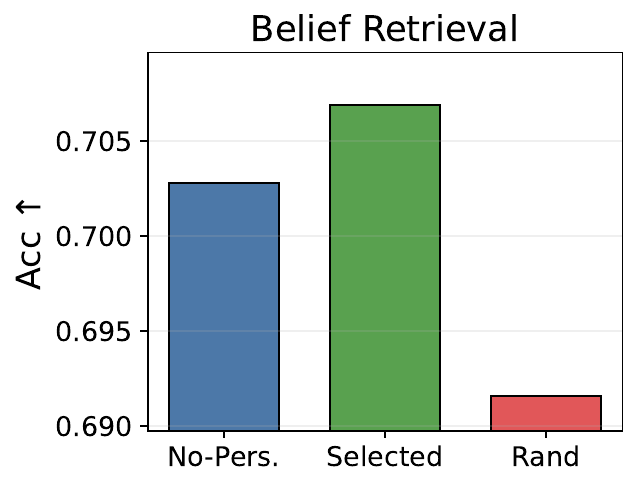}
\end{subfigure}
\begin{subfigure}[t]{0.24\textwidth}
\centering
\includegraphics[width=\linewidth]{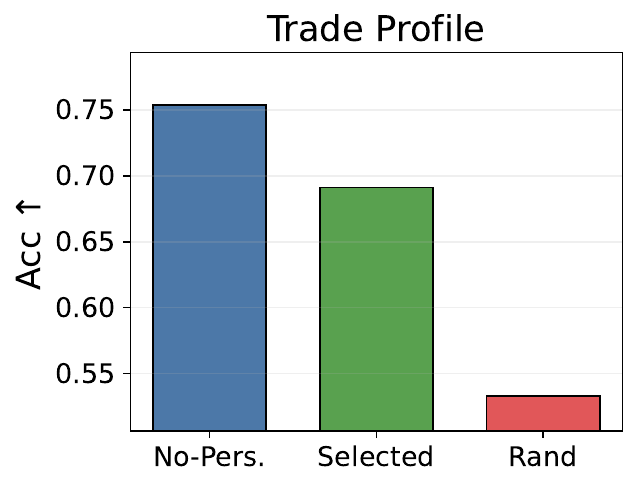}
\end{subfigure}
\begin{subfigure}[t]{0.24\textwidth}
\centering
\includegraphics[width=\linewidth]{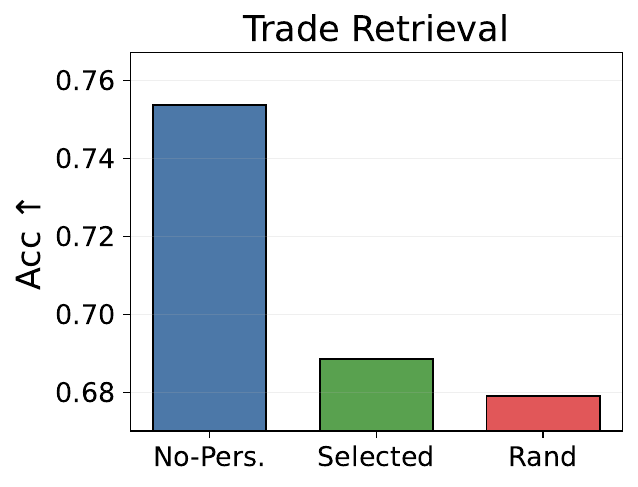}
\end{subfigure}
\caption{Ablation results for direction and choice accuracy.}
\label{fig:ablation_accuracy}
\end{figure*}

Figure~\ref{fig:ablation_accuracy} studies profile and retrieval design choices for choice and direction accuracy. For Belief, selected profiles improve accuracy over the unaugmented condition, while random profiles perform worse; the same pattern appears for Belief retrieval, though the gain is smaller. The absolute-error ablations in Figure~\ref{fig:ablation_error} show an even clearer effect: selected profiles substantially reduce Belief error, while random profiles remain worse. Thus, Belief improvements are not merely caused by adding more text to the prompt; they depend on behaviorally relevant profile information.

For Trade, the ablations are more cautionary. Selected profiles reduce direction accuracy relative to the direct condition, but still outperform random profiles by a large margin. Retrieval shows a similar pattern: selected support examples are better than random support examples, yet both fall below the direct-history condition. Amount error is especially sensitive to irrelevant profile information, with random profiles producing very large errors. These findings reinforce the main result: transaction-level prediction requires fine-grained context, and poorly matched or overly compressed history can actively hurt.

\paragraph{Oracle and error complementarity.}

\begin{figure*}[t]
\centering
\begin{subfigure}[t]{0.45\textwidth}
\centering
\includegraphics[width=\linewidth]{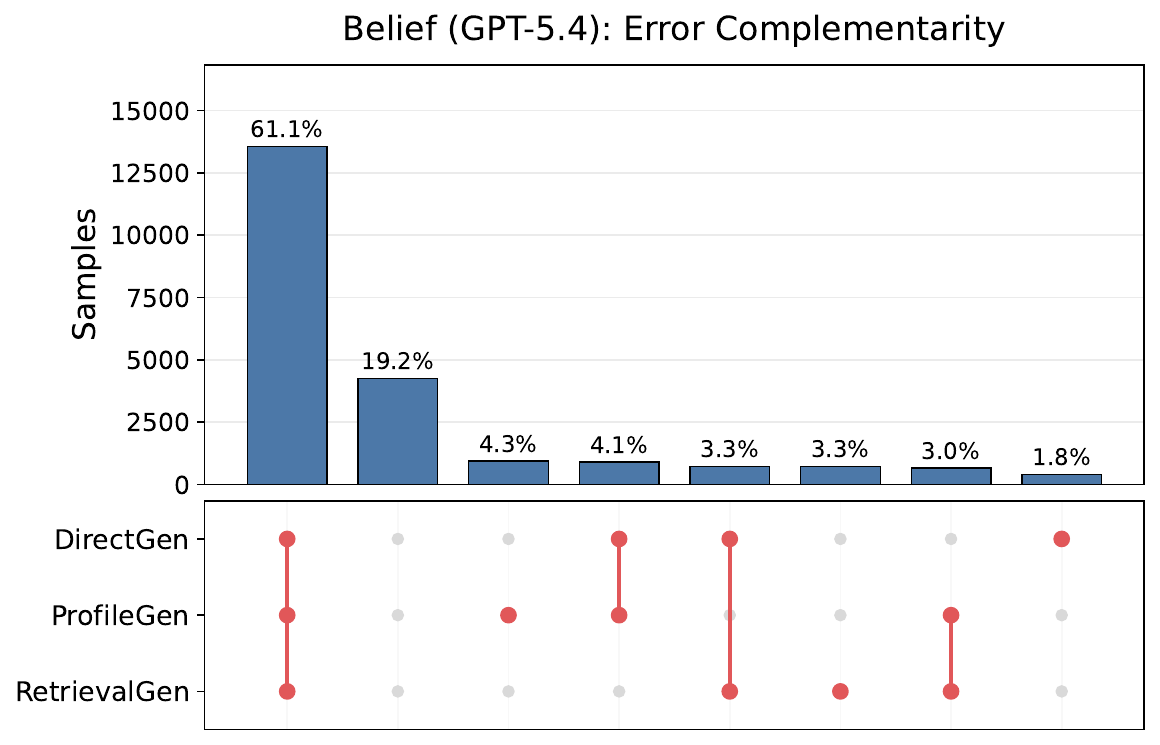}
\end{subfigure}
\begin{subfigure}[t]{0.45\textwidth}
\centering
\includegraphics[width=\linewidth]{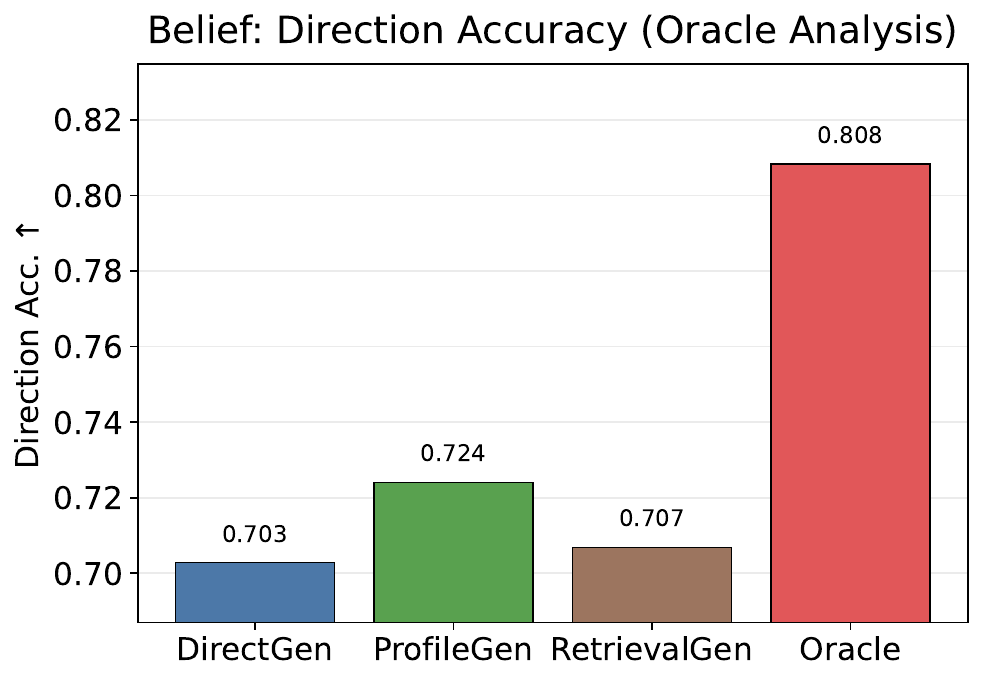}
\end{subfigure}
\caption{Selected oracle and error-complementarity diagnostics for Belief-layer decision structure.}
\label{fig:belief_oracle_main}
\end{figure*}

Figure~\ref{fig:belief_oracle_main} shows selected oracle and upset diagnostics for the Belief layer. Oracle and upset analyses are included to diagnose remaining headroom and error complementarity. Oracle experiments test whether additional structured signal can raise performance beyond the main input interfaces, while upset plots measure whether different models and history interfaces fail on the same instances or on complementary subsets. We include the full set of oracle and upset plots in the appendix.

The upset plots show that most errors are shared across history interfaces rather than being isolated to a single method. Both Belief and Trade contain a substantial shared-error region, while method-specific error regions remain nontrivial. This indicates that the benchmark contains a hard core of instances that current prompting interfaces do not solve, while still leaving room for method complementarity through ensembles or better history selection.

\section{Related Work}

Personalization benchmarks often evaluate persona consistency, profile retrieval, personalized response generation, or preference following from textual personas and dialogue histories \citep{zhang2018personalizing,zheng2019personalized,salemi2024lamp,jiang2025know,zhang2026preferencealignment}. Recent work extends this direction to constructed preferences and implicit user information \citep{tan2025personabench,tao2025personafeedback,zhao2025llms,wu2026knowme,liang2026benchmarking}. \textsc{BehaviorBench} differs by deriving labels from revealed behavior in public decision traces: systems must infer stance, confidence, direction, and amount from observed histories rather than explicit persona statements.

Real-user lifelog and memory benchmarks ground personalization in longitudinal evidence, but typically evaluate retrieval or question answering over personal archives \citep{tran2022llqa,tran2025openlifelogqa,gurrin2022lsc,zhang2026memorycd}. Generative agents and user simulators provide controllable behavioral trajectories \citep{park2023generative,li2025behaviorchain,zhang2024usimagent,zhu2026realusersim}, but synthetic behavior may diverge from real users \citep{hu2025simbench,lee2025realtalk,zhou2026sim2real}. \textsc{BehaviorBench} uses real public behavioral traces as the source of users and labels; generated summaries appear only as one evaluated input representation.

\section{Limitations and Broader Impact}

\textsc{BehaviorBench} is based on observed prediction-market behavior and should not be generalized to all human decision-making. Wallets are pseudonymous behavioral accounts, not verified people: one actor may control multiple wallets, a wallet may reflect multiple actors, and some activity may be automated or coordinated. The labels capture revealed behavior rather than private beliefs, and trade amount prediction can depend on portfolio constraints, liquidity, fees, or strategy that are not fully observable. The same user-modeling capability studied here could also be misused for targeting, manipulation, or surveillance. We therefore frame the dataset as an evaluation resource, do not attempt identity inference or off-chain linkage, and provide license, data-card, and use-policy documentation with the release.

\section{Conclusion}

We introduced \textsc{BehaviorBench}, a benchmark for modeling user decisions from real-world behavioral traces. By separating Belief prediction from Trade prediction, and by comparing no-history, direct-history, profile-based, and retrieval-based interfaces, the benchmark makes history representation part of the evaluation. Current generative models extract useful behavioral signal, but performance remains limited and depends strongly on task layer, metric, and history representation. \textsc{BehaviorBench} provides a basis for evaluating personalized methods that reason over heterogeneous real-world user behavior over time.

\bibliographystyle{plainnat}
\bibliography{references}

\newpage
\appendix

\section{Dataset Documentation and Intended Use}
\label{app:dataset_documentation}

\paragraph{Intended use.}
\textsc{BehaviorBench} is designed for evaluating systems that infer personalized decisions from historical behavioral traces. Appropriate uses include benchmarking history representations, retrieval strategies, profile generation methods, memory mechanisms, and calibration methods for revealed decision prediction. The benchmark should not be used to identify wallet owners, infer off-chain identities, target individuals, or provide financial advice.

The benchmark is most suitable for research questions where the object of study is the interface between behavioral evidence, memory, and prediction. For example, it can be used to ask whether a system benefits more from recent raw history, compressed profiles, or retrieved peer evidence; whether final-position prediction and transaction prediction require different context; or whether calibration metrics expose failures that are hidden by accuracy. It is not intended as a simulator of a complete market, a trading strategy backtest, or a measure of individual rationality.

\paragraph{Artifact contents.}
The released artifact contains benchmark-ready Belief and Trade splits, support-wallet pools for retrieval experiments, input-construction and evaluation scripts, and metadata documenting data provenance and preprocessing. The benchmark files are released separately from the raw chain archive when possible, so users can reproduce evaluation without re-querying external services. We document that bit-level reconstruction from raw public APIs and chain nodes may depend on the historical API snapshot, node behavior, and software versions.

The benchmark-ready files are organized around instance-level evaluation rather than raw chain replay. Each Belief instance contains the target market context, the target wallet identifier used internally for split construction, the visible history representation for the relevant method, and the held-out final side and confidence label. Each Trade instance contains the target transaction context, visible history, and held-out direction and amount labels. Retrieval experiments additionally consume support-pool files that are disjoint from the target-wallet test set.

\paragraph{Reproducibility path.}
There are two intended levels of reproducibility. The first reproduces the benchmark evaluation from released processed files: users load the benchmark splits, run the input-construction or model-evaluation scripts, parse structured outputs, and compute the reported metrics. This is the primary reproducibility path and does not require re-querying public APIs. The second reconstructs the processed benchmark from public sources using the preprocessing scripts. This path is useful for auditing provenance, but exact bit-level equality can depend on external API snapshots, chain-node behavior, and software-version locking.

\paragraph{Croissant and RAI fields.}
The dataset release is hosted on a platform that supports Croissant metadata generation for datasets. We additionally provide Responsible AI documentation fields describing data source provenance, public/pseudonymous identifiers, intended use, out-of-scope use, privacy considerations, preprocessing filters, and known limitations. These fields are included to align the artifact with the NeurIPS Evaluations \& Datasets track expectation that benchmark submissions be usable as standalone scientific resources.

\paragraph{Privacy and consent.}
The source data consist of public on-chain logs and public market metadata. Wallet addresses are public pseudonymous identifiers rather than verified real-world identities. We do not attempt identity inference, link wallets to off-chain identities, or release enriched identity attributes. Because the benchmark is reconstructed from public behavioral traces rather than collected through an intervention or crowdsourcing study, there are no recruited participants or compensation procedures.

\paragraph{Known caveats.}
Wallets are behavioral accounts, not verified people. A wallet may be controlled by multiple actors, one actor may control multiple wallets, and some wallets may reflect automated or coordinated strategies. Labels should therefore be interpreted as revealed behavioral targets attached to pseudonymous accounts, not as ground-truth psychological beliefs. These caveats motivate our focus on benchmark evaluation rather than individual-level interpretation.

\section{Reproducibility Checklist for Benchmark Users}
\label{app:reproducibility_path}

This section summarizes the practical reproduction path expected for benchmark users. The paper's primary empirical claims depend on the released benchmark splits, input-construction templates, model outputs or generation scripts, parsing code, and metric computation. Users who want to reproduce the benchmark from raw public sources can additionally run the raw-data reconstruction pipeline, but that path is more sensitive to external-service versioning.

\paragraph{Processed benchmark evaluation.}
The recommended reproduction path starts from the released benchmark directory. A user loads the Belief or Trade split files, chooses a history interface, generates model or agent outputs with the corresponding input template, parses outputs into the structured prediction schema, and computes the reported metrics. Belief evaluation reports choice accuracy and confidence absolute error on a $[0,1]$ scale. Trade evaluation reports BUY/SELL direction accuracy and amount absolute error. Retrieval-based methods additionally load the disjoint support pool and construct retrieved examples before prediction.

\subsection{Metric Formulas}
\label{app:metric_formulas}

For a set of $N$ Belief instances, Choice Accuracy is
\[
\mathrm{Acc}_{\mathrm{belief}}=\frac{1}{N}\sum_{j=1}^N
\mathbf{1}\{\hat{y}^{\mathrm{side}}_j=y^{\mathrm{side}}_j\},
\]
and Confidence MAE is
\[
\mathrm{MAE}_{\mathrm{conf}}=\frac{1}{N}\sum_{j=1}^N
\left|\hat{y}^{\mathrm{conf}}_j-y^{\mathrm{conf}}_j\right|.
\]
For a set of $N$ Trade instances, Direction Accuracy is
\[
\mathrm{Acc}_{\mathrm{trade}}=\frac{1}{N}\sum_{j=1}^N
\mathbf{1}\{\hat{y}^{\mathrm{dir}}_j=y^{\mathrm{dir}}_j\},
\]
and amount error is reported as Median Absolute Error,
\[
\mathrm{MedAE}_{\mathrm{amt}}=
\operatorname{median}_{j\in\{1,\ldots,N\}}
\left|\hat{y}^{\mathrm{amt}}_j-y^{\mathrm{amt}}_j\right|.
\]

\paragraph{Raw reconstruction audit.}
For auditing, the preprocessing scripts reconstruct wallet--market--outcome trajectories from public metadata and on-chain logs. The raw pipeline collects market/event metadata from public APIs, expands wallet candidates from \texttt{topic\_addresses}, reconstructs signed position changes from BUY/SELL directions and parsed amounts, and applies the filtering chain in Appendix~\ref{app:cohort_filtering}. Exact bit-level agreement with the released files may depend on API snapshots, chain-node responses, and software versions; therefore, the processed benchmark files are the authoritative evaluation artifact.

\paragraph{Failure handling.}
Malformed outputs are handled consistently within each task layer. Belief parsing uses fallback behavior so that malformed generations do not abort an evaluation run. Trade parsing marks invalid or unparsable outputs as failed predictions, which are counted in the evaluation. This distinction should be preserved by downstream users because parse failures are part of the measured robustness of transaction-level prediction.

\section{Input and Output Schemas}
\label{app:prompt_schema}

This section describes the input and output interfaces at the level needed to reproduce the benchmark. The released code contains the exact templates used for the generative-model experiments; here we summarize the fields, history placement, and output constraints. Agent systems can consume the same fields through a memory module, retrieval module, or tool call rather than through a single prompt. All interfaces share the same target instance, but differ in whether and how they expose behavioral history.

\paragraph{No Personalization.}
The no-personalization input contains only the target event or transaction context. It does not include wallet history, profile summaries, retrieved examples, wallet identifiers as behavioral evidence, or target-wallet labels. This baseline isolates what can be inferred from the event context alone.

\begin{promptbox}
System: Predict the user's decision for the target instance.
Input:
  TargetContext: {event/market or transaction context}
Output:
  Return only valid JSON matching the task schema.
\end{promptbox}

\paragraph{DirectGen.}
DirectGen adds a recency-truncated target-wallet history. For Belief, the default maximum history length is 30 events. For Trade, the benchmark uses the windowed \texttt{HistoryJSON} field. The input does not provide the held-out target label.

\begin{promptbox}
System: Predict the target user's decision from the target context
and the user's prior behavior.
Input:
  TargetContext: {...}
  HistoryJSON: [most recent visible history items]
Output:
  Return only valid JSON matching the task schema.
\end{promptbox}

\paragraph{ProfileGen.}
ProfileGen replaces raw target-wallet history with a structured profile generated from pre-target visible history. The profile schema contains \texttt{topic\_preferences}, \texttt{decision\_style}, \texttt{confidence\_pattern}, \texttt{consistency}, and \texttt{notes}. The profile generator never observes target-instance labels.

\begin{promptbox}
System: Predict the target user's decision from the target context
and the user's historical profile.
Input:
  TargetContext: {...}
  Profile:
    topic_preferences: ...
    decision_style: ...
    confidence_pattern: ...
    consistency: ...
    notes: ...
Output:
  Return only valid JSON matching the task schema.
\end{promptbox}

\paragraph{RetrievalGen.}
RetrievalGen augments the target with retrieved support-wallet evidence. Retrieved wallets come from a disjoint support pool and exclude the target wallet. Belief retrieval uses structured behavior-signature overlap, while Trade retrieval uses a market-first structured similarity pipeline with top-$k=5$ support examples in the main generation setting.

\begin{promptbox}
System: Predict the target user's decision from the target context
and behaviorally similar support examples.
Input:
  TargetContext: {...}
  RetrievedExamples: [
    {support_context: ..., support_behavior: ..., support_label: ...}
  ]
Output:
  Return only valid JSON matching the task schema.
\end{promptbox}

\paragraph{Belief output schema.}
Belief predictions are parsed as structured outputs containing a final side and confidence. Confidence is interpreted on the $[0,1]$ scale. Invalid or malformed Belief outputs are handled by the benchmark fallback parser.

\begin{promptbox}
{
  "final_side": "YES" | "NO",
  "confidence": number
}
\end{promptbox}

\paragraph{Trade output schema.}
Trade predictions are parsed as structured outputs containing direction and amount. Direction is evaluated as BUY/SELL accuracy. Amount is evaluated with absolute-error metrics. Invalid, missing, or unparsable Trade fields are marked as failed predictions.

\begin{promptbox}
{
  "direction": "BUY" | "SELL",
  "amount": number
}
\end{promptbox}

\section{Example Benchmark Instances}
\label{app:example_instances}

This section gives shortened, anonymized examples of the record format used by the released benchmark files. The examples preserve the benchmark field structure, but truncate histories, retrieved evidence, and free-text event metadata for readability. Wallet identifiers are replaced with stable pseudonyms, and the examples should be read as documentation of the input/output contract rather than as a release of additional identifying attributes.

\paragraph{Belief DirectGen instance.}
A Belief instance asks for the final revealed side and confidence of a target wallet on a target market. DirectGen exposes a recency-truncated target-wallet history before the target instance.

\begin{promptbox}
{
  "task": "belief",
  "interface": "direct",
  "wallet": "wallet_A17",
  "target_context": {
    "event": {
      "title": "Federal Reserve rate decision",
      "question": "Will the Federal Reserve cut interest rates at
                   the next FOMC meeting?",
      "category": "macroeconomics",
      "deadline": "YYYY-MM-DD"
    },
    "outcomes": ["YES", "NO"],
    "resolution_status": "resolved"
  },
  "history": [
    {"event": "Will annual inflation be above the reported threshold?",
     "category": "macroeconomics", "final_side": "YES",
     "confidence": 0.74},
    {"event": "Will a named candidate win the national election?",
     "category": "politics", "final_side": "NO",
     "confidence": 0.62}
  ],
  "label": {"final_side": "YES", "confidence": 0.81}
}
\end{promptbox}

\paragraph{ProfileGen excerpt.}
ProfileGen replaces raw target-wallet history with a structured profile generated only from pre-target visible behavior. The profile is used as input evidence; the held-out target label is not used when constructing it.

\begin{promptbox}
{
  "task": "belief",
  "interface": "profile",
  "wallet": "wallet_A17",
  "target_context": {
    "event": "Will the Federal Reserve cut interest rates at the
              next FOMC meeting?",
    "outcomes": ["YES", "NO"]
  },
  "profile": {
    "topic_preferences": ["politics", "macro events"],
    "decision_style": "often takes directional positions early",
    "confidence_pattern": "higher confidence on repeated topics",
    "consistency": "usually maintains the same final side",
    "notes": "generated from visible pre-target history only"
  },
  "label": {"final_side": "YES", "confidence": 0.81}
}
\end{promptbox}

\paragraph{RetrievalGen excerpt.}
RetrievalGen adds support-wallet evidence retrieved from a disjoint pool. In the released Belief retrieval files, retrieval is time-causal: support rows must precede the target stance block, and the target wallet is excluded.

\begin{promptbox}
{
  "task": "belief",
  "interface": "retrieval",
  "wallet": "wallet_A17",
  "target_context": {
    "event": "Will the Federal Reserve cut interest rates at the
              next FOMC meeting?",
    "outcomes": ["YES", "NO"]
  },
  "retrieved_examples": [
    {"support_wallet": "support_wallet_083",
     "support_event": "Federal Reserve rate decision",
     "support_behavior": {"final_side": "YES", "confidence": 0.69}},
    {"support_wallet": "support_wallet_219",
     "support_event": "Federal Reserve rate decision",
     "support_behavior": {"final_side": "NO", "confidence": 0.57}}
  ],
  "label": {"final_side": "YES", "confidence": 0.81}
}
\end{promptbox}

\paragraph{Trade DirectGen instance.}
A Trade instance asks for the next transaction direction and amount. The visible history is a windowed transaction history rather than the wallet's unbounded raw history.

\begin{promptbox}
{
  "task": "trade",
  "interface": "direct",
  "wallet": "wallet_B42",
  "target_context": {
    "event": {
      "title": "Monthly inflation release",
      "question": "Will the next CPI print be above the stated
                   threshold?",
      "category": "macroeconomics"
    },
    "outcome": "YES",
    "block": 79881234,
    "position_before": 120
  },
  "history": [
    {"event": "Monthly inflation release", "outcome": "YES",
     "direction": "BUY", "amount": 100, "block": 79870001},
    {"event": "Monthly inflation release", "outcome": "YES",
     "direction": "SELL", "amount": 40, "block": 79876010}
  ],
  "label": {"direction": "BUY", "amount": 75,
            "action_label": "accumulate"}
}
\end{promptbox}

\section{Compute and Asset Release Notes}
\label{app:compute_release}

\paragraph{Compute profile.}
The benchmark does not train new neural models. Dataset construction and metric computation are CPU-oriented preprocessing and evaluation jobs over tabular/parquet files. In our reported instantiation, the main computational cost comes from generative-model inference for the evaluated history interfaces. Closed-model experiments are run through hosted chat-completion APIs; open-weight experiments can be reproduced either through hosted inference endpoints or local inference infrastructure sufficient for the selected model size. Future agent implementations may add retrieval, memory-update, or tool-use overhead, but the benchmark input/output contract remains unchanged.

\paragraph{Experiment scale.}
Main experiments and ablations are run on the full test set, while history-scaling experiments use a 100-wallet subset to keep repeated history-length sweeps tractable. Exact wall-clock time and monetary cost depend on API provider throughput, rate limits, batching, retry policy, and the chosen set of models. The released evaluation scripts expose the number of instances per split and method so users can estimate token and inference cost before running a full benchmark.

\paragraph{License and third-party assets.}
The dataset is derived from public market metadata and public on-chain logs. The processed benchmark files and accompanying materials are released under Apache-2.0, with a dataset card documenting provenance, intended use, privacy considerations, and upstream source terms. Downstream users should follow the dataset card and respect any upstream terms-of-use constraints for public API metadata and chain logs.

\section{Cohort Construction and Filtering}
\label{app:cohort_filtering}

Table~\ref{tab:cohort_filtering_chain} gives the full wallet-cohort construction and filtering chain used in the main benchmark pipeline. These thresholds are quality-control criteria for constructing a stable evaluation cohort from public behavioral traces; they should not be interpreted as claims that filtered wallets are necessarily invalid users.

The filtering pipeline has three goals. First, it removes wallets with too little activity to support personalized evaluation. Second, it removes extreme outliers whose trajectories are likely to dominate statistics or exceed practical context budgets. Third, it reduces artifacts that arise when raw logs are converted into position trajectories, such as impossible negative-position paths or highly concentrated burst activity. We therefore treat the thresholds as benchmark-construction choices rather than behavioral validity judgments.

The common cohort is formed before the final V2 filters so that Belief and Trade experiments start from the same broad population of repeatedly active wallets. The later filters then specialize this population into stable, temporally standardized benchmark instances. The \texttt{pre80m} cutoff is especially important for preventing accidental mixing of experimental vintages, since all main V2 results use the same block cutoff. Substantively, this cutoff keeps the main benchmark within the data vintage ending in December 2025. We use it to avoid blending the pre-2026 user-behavior regime with the sharp transaction-volume increase we observe beginning in January 2026, which is plausibly associated with more automated or agent-mediated activity. We do not treat the cutoff as a claim that every later transaction is non-human; rather, it is a benchmark-construction choice that reduces a visible regime shift in the raw data.

\begin{table*}[t]
\centering
\footnotesize
\setlength{\tabcolsep}{3pt}
\caption{Wallet-cohort construction and filtering chain for the main benchmark pipeline.}
\label{tab:cohort_filtering_chain}
\begin{tabular}{L{0.23\textwidth} L{0.22\textwidth} L{0.43\textwidth}}
\toprule
Stage & Criterion & Purpose \\
\midrule
\multirow{4}{=}{Raw on-chain logs to candidate pool}
& $\texttt{logs} \geq 30$ & Retain wallets with sufficient observed activity. \\
& $\texttt{logs} \leq 5{,}000{,}000$ & Remove extreme log-volume outliers. \\
& $\texttt{txs} \geq 20$ & Require at least 20 distinct transactions. \\
& $\texttt{contracts} \geq 2$ & Remove wallets interacting with only one contract. \\
\midrule
Candidate pool to common cohort
& $\texttt{market\_count} \geq 10$ & Form a common pre-\texttt{mk10plus} cohort of 31,319 wallets with at least 10 interacted markets. \\
\midrule
\multirow{2}{=}{Consistency and temporal window}
& \texttt{nonneg clean} & Remove inconsistent negative-position trajectories. \\
& \texttt{pre80m}: $\texttt{block} \leq 80{,}000{,}000$ & Standardize the V2 temporal window, keeping the main benchmark in the data vintage ending in December 2025 and before the January 2026 transaction-volume surge. \\
\midrule
\multirow{2}{=}{User--market support}
& \texttt{mk3\_noclose}: $\texttt{user} \times \texttt{market} \geq 3$ & Require repeated observations within each user--market sequence. \\
& closing trade not required & Avoid discarding sequences without an explicit close. \\
\midrule
\multirow{4}{=}{Normal users v1.1}
& $\texttt{rows/user} \in [50, 10000]$ & Remove very sparse and extremely large wallet histories. \\
& $\texttt{markets/user} \geq 5$ & Require behavior across multiple markets. \\
& $\texttt{max\_market\_share} \leq 0.8$ & Remove wallets dominated by a single market. \\
& $\texttt{rows/user} \times \texttt{market} \leq 300$ & Remove extreme user--market sequence lengths before final capping. \\
\midrule
\multirow{3}{=}{Realness v1.2}
& $\texttt{max\_block\_share} \leq 0.3$ & Remove block-level activity concentration. \\
& $\texttt{max\_tx\_share} \leq 0.3$ & Remove transaction-level activity concentration. \\
& $\texttt{burst\_block\_ratio} \leq 0.6$ & Remove highly bursty activity patterns. \\
\midrule
Final cap
& \texttt{cap30}: $\texttt{rows/user} \times \texttt{market} \leq 30$ & Limit each user--market sequence to at most 30 rows. \\
\bottomrule
\end{tabular}
\end{table*}
\FloatBarrier

\section{Additional Decomposed Results}
\label{app:decomposed_results}

\begin{figure}[H]
\centering
\begin{subfigure}[t]{0.8\textwidth}
\centering
\includegraphics[width=\linewidth]{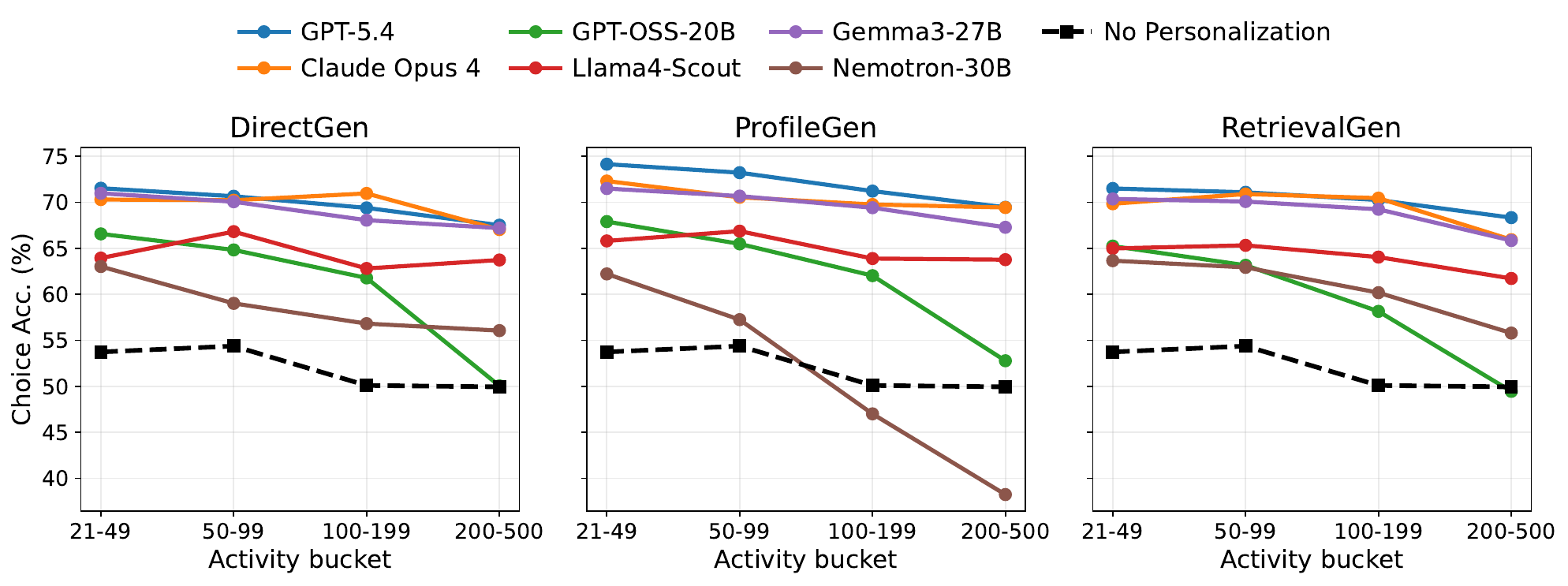}
\end{subfigure}
\begin{subfigure}[t]{0.8\textwidth}
\centering
\includegraphics[width=\linewidth]{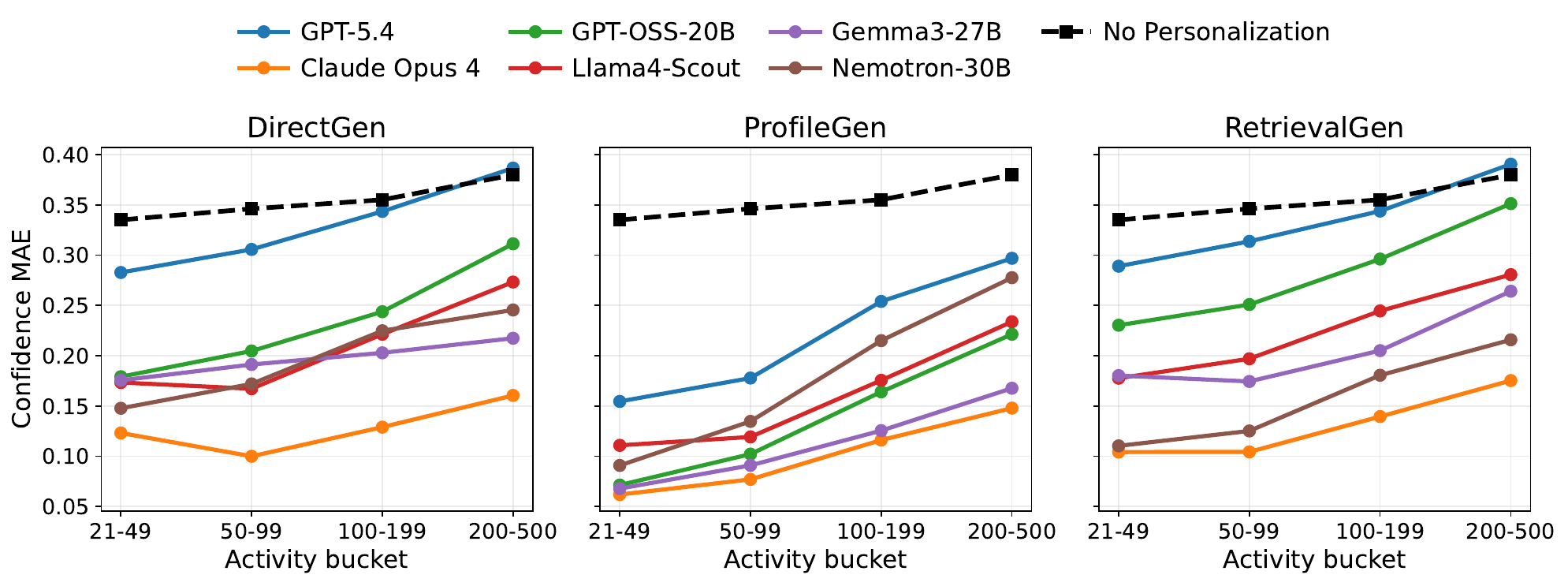}
\end{subfigure}
\caption{
Decomposed results of the Belief layer.
}
\label{fig:decomposed_belief}
\end{figure}

\begin{figure}[H]
\centering
\begin{subfigure}[t]{0.8\textwidth}
\centering
\includegraphics[width=\linewidth]{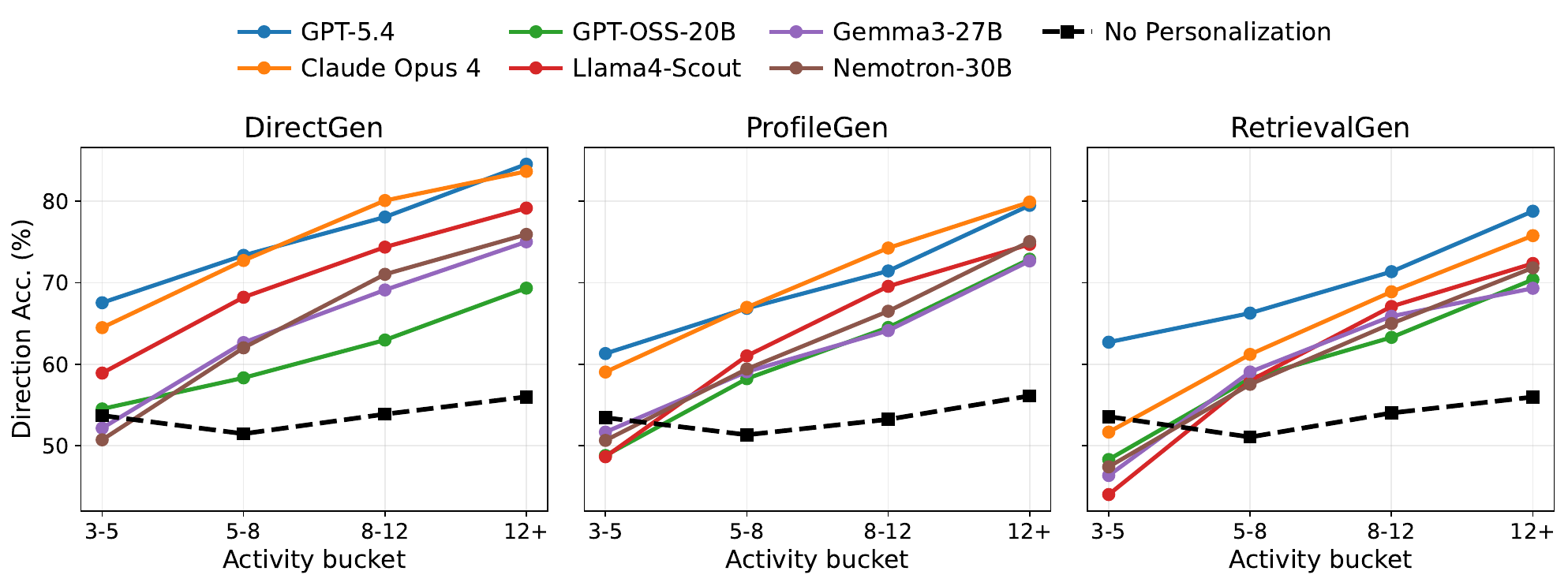}
\end{subfigure}
\begin{subfigure}[t]{0.8\textwidth}
\centering
\includegraphics[width=\linewidth]{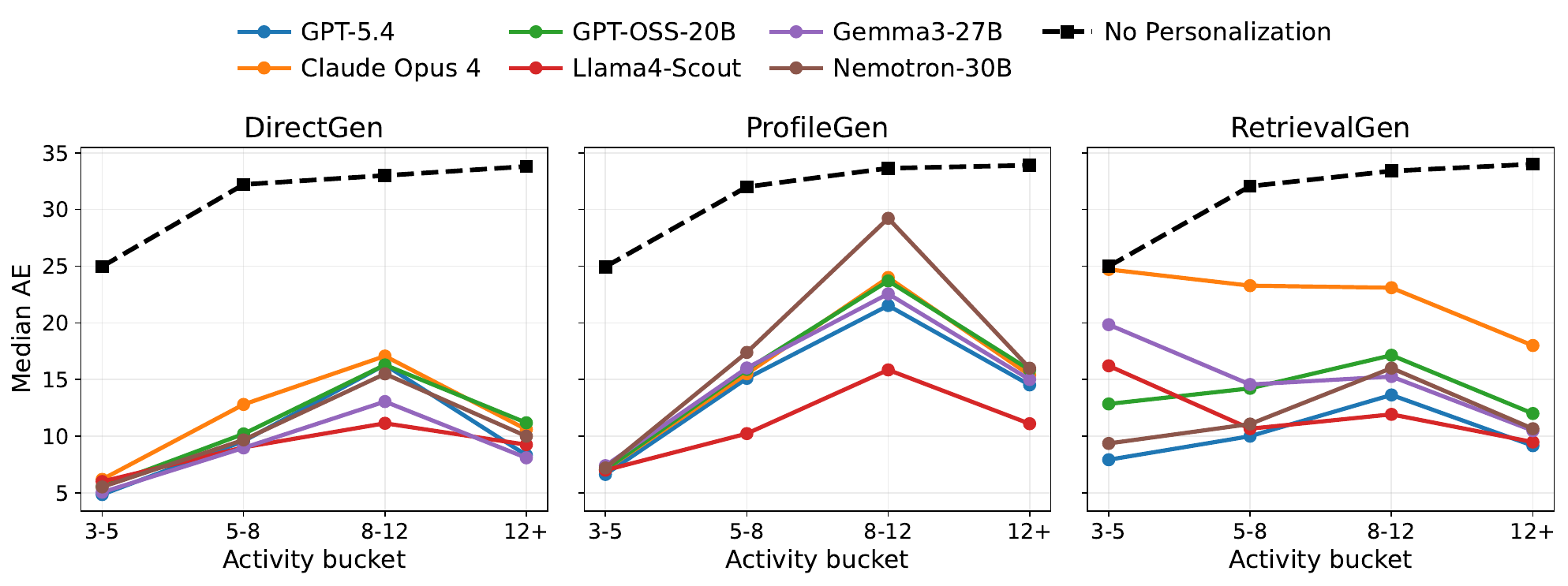}
\end{subfigure}
\caption{
Decomposed results of the Trade layer.
}
\label{fig:decomposed_trade}
\end{figure}

Figures~\ref{fig:decomposed_belief} and~\ref{fig:decomposed_trade} break down selected main results by user-activity bucket. Belief buckets are defined by the number of markets a wallet interacted with, while Trade buckets are defined by average transactions per market. These plots support the main-paper claim that user activity affects the two task layers differently.

The decomposition is included because aggregate performance can hide qualitatively different sources of difficulty. A wallet with many interacted markets provides broad evidence about topical preferences, but it may also represent a more heterogeneous decision maker. Conversely, a wallet with many transactions per market provides repeated local evidence about how the wallet adjusts exposure within a specific market. The two bucket definitions therefore intentionally differ: Belief uses market breadth, while Trade uses within-market density.

For Belief, the decomposed plots should be read as a stress test of generalization across increasingly broad user histories. Declining accuracy in higher market-count buckets suggests that more history is not automatically easier to exploit; broad histories can introduce topic diversity, inconsistent stances, or context that is difficult to compress into a compact memory or input representation. The accompanying error plot shows whether this difficulty is only about choosing the correct side or also about calibrating the final confidence.

For Trade, the decomposed plots focus on action density. Direction accuracy generally improves as average transactions per market increases, which indicates that repeated within-market behavior gives models more usable local signal. Amount error is less monotonic, suggesting that predicting magnitude depends on additional factors such as exposure, partial closing behavior, and market-specific liquidity signals that are only partially captured by the benchmark input.

\section{Additional Ablations}
\label{app:additional_ablations}

\begin{figure}[H]
\centering
\begin{subfigure}[t]{0.24\textwidth}
\centering
\includegraphics[width=\linewidth]{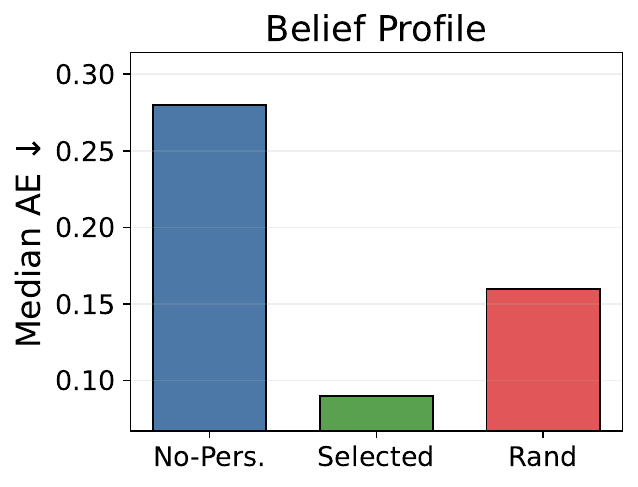}
\end{subfigure}
\begin{subfigure}[t]{0.24\textwidth}
\centering
\includegraphics[width=\linewidth]{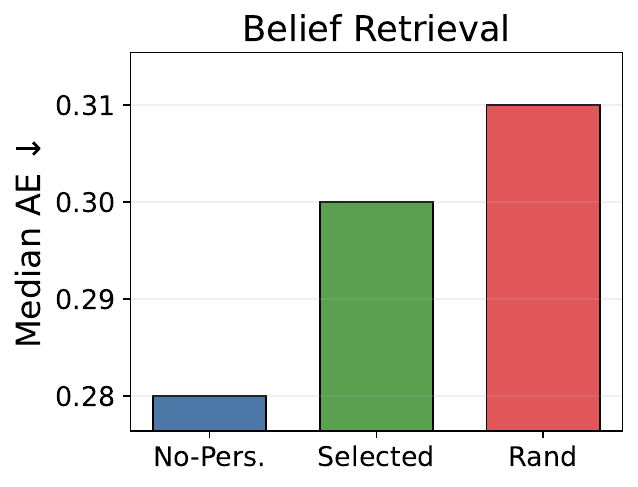}
\end{subfigure}
\begin{subfigure}[t]{0.24\textwidth}
\centering
\includegraphics[width=\linewidth]{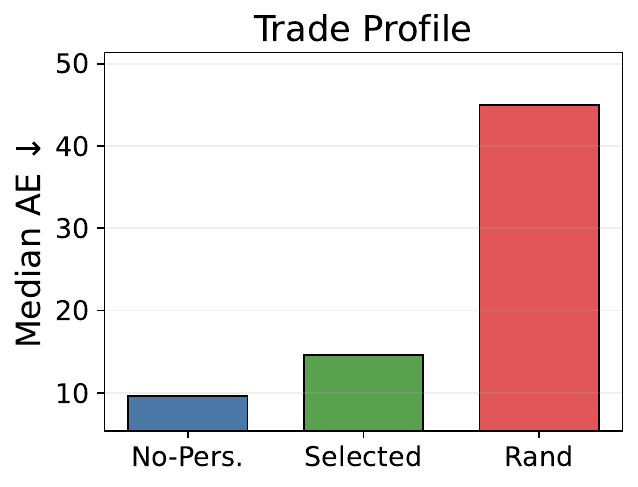}
\end{subfigure}
\begin{subfigure}[t]{0.24\textwidth}
\centering
\includegraphics[width=\linewidth]{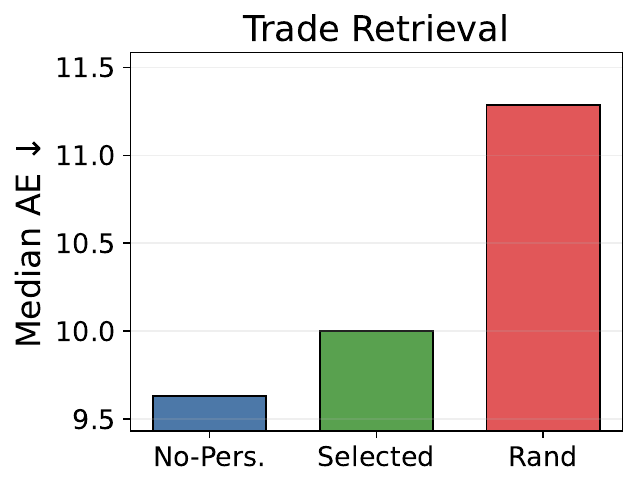}
\end{subfigure}
\caption{
Ablation results for absolute-error metrics.
}
\label{fig:ablation_error}
\end{figure}

Figure~\ref{fig:ablation_error} complements the accuracy ablations in the main paper by reporting absolute-error metrics. These plots test whether selected profiles or selected retrieved examples improve calibration and amount prediction, not only side or direction prediction.

The ablations compare behaviorally matched context against deliberately weaker context. For profile experiments, the selected condition uses a profile generated from the target wallet's visible historical behavior, while the random condition substitutes an unrelated profile. For retrieval experiments, the selected condition uses the structured retrieval pipeline, while the random condition injects support examples without behavioral matching. These controls test whether gains come from relevant personalization rather than simply from adding more text or examples to the input.

The Belief absolute-error results are particularly informative because the confidence target is continuous. A method can predict the correct side while still being poorly calibrated about the strength of the final revealed position. Selected profiles substantially reduce Belief error relative to unaugmented and random-profile conditions, which supports the interpretation that profile summaries capture stable user-level tendencies useful for calibration.

The Trade error ablations are more fragile. Random or mismatched profile information can produce large amount errors, and selected retrieval does not consistently match direct-history performance. This pattern is consistent with the main paper's conclusion that transaction-level amount prediction is sensitive to local sequence context and can be harmed by irrelevant or overly compressed behavioral evidence.

\section{Oracle Experiments}
\label{app:oracle_experiments}

\begin{figure}[H]
\centering
\begin{subfigure}[t]{0.24\textwidth}
\centering
\includegraphics[width=\linewidth]{pics/oracle/belief_direction_oracle_v2.pdf}
\end{subfigure}
\begin{subfigure}[t]{0.24\textwidth}
\centering
\includegraphics[width=\linewidth]{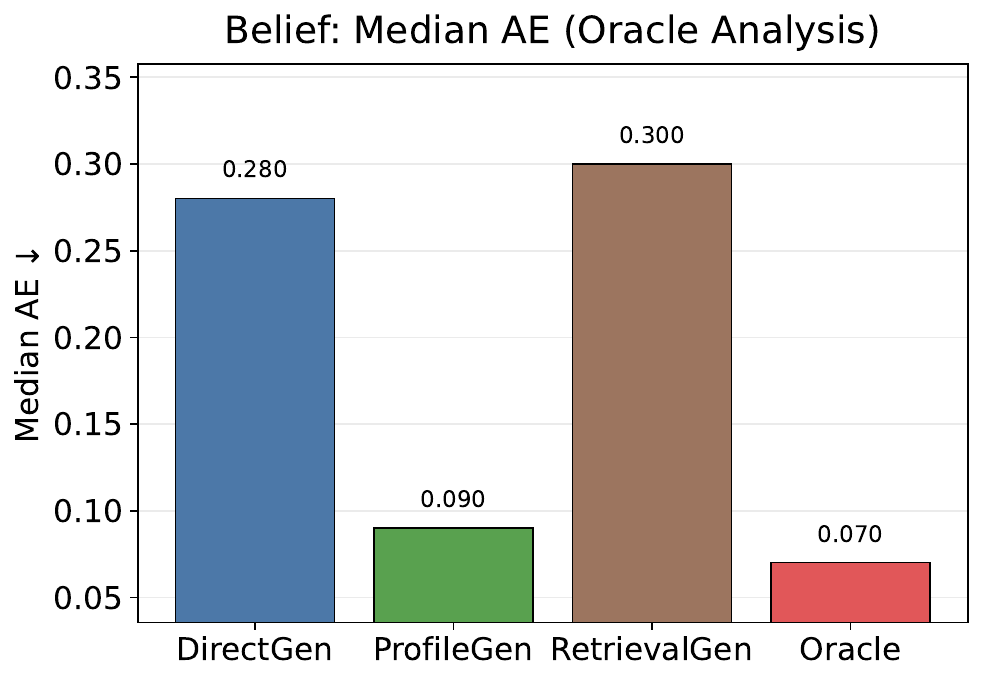}
\end{subfigure}
\begin{subfigure}[t]{0.24\textwidth}
\centering
\includegraphics[width=\linewidth]{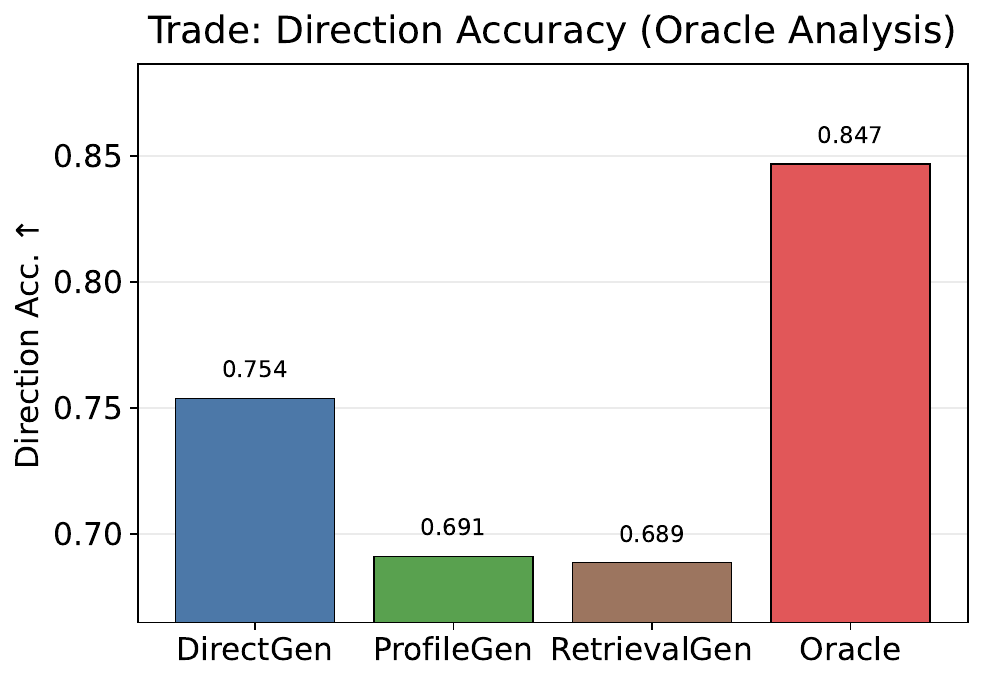}
\end{subfigure}
\begin{subfigure}[t]{0.24\textwidth}
\centering
\includegraphics[width=\linewidth]{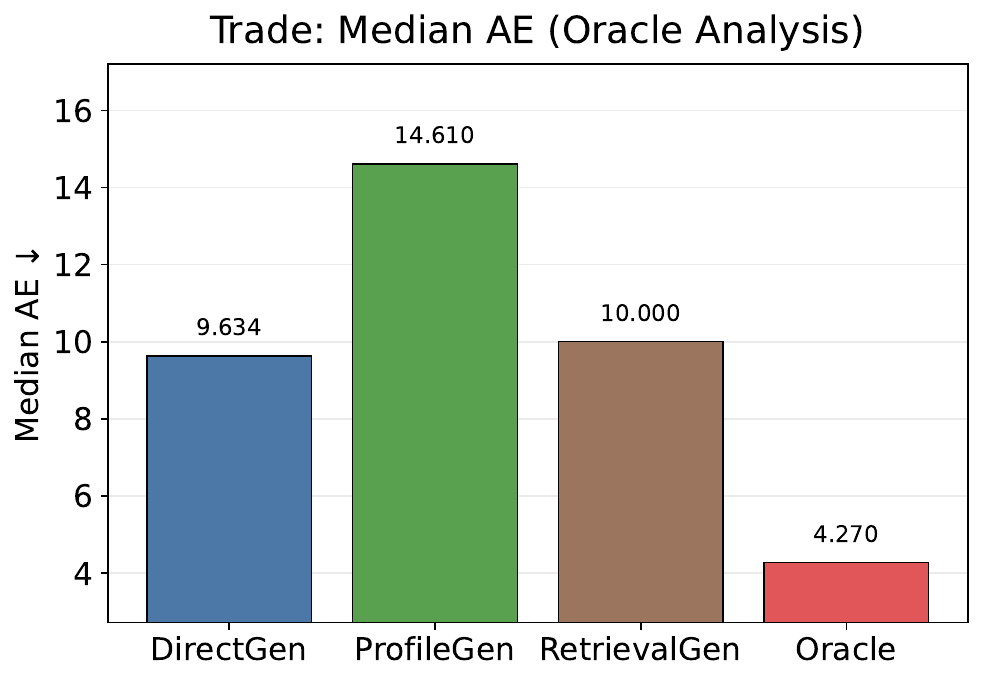}
\end{subfigure}

\caption{
Oracle experiments across both layers and metrics.
}
\label{fig:oracle_all}
\end{figure}

Figure~\ref{fig:oracle_all} reports oracle-style diagnostics across both task layers and metrics. These experiments are intended to estimate remaining headroom under stronger structured signal than the main input interfaces provide.

The oracle experiments are not proposed as deployable methods. Instead, they serve as diagnostics for the benchmark. If an oracle-style signal substantially improves performance, then the corresponding task contains exploitable structure that current input interfaces do not fully expose. If oracle performance remains limited, then the bottleneck may lie in missing information, label ambiguity, or intrinsically hard instances rather than in the choice of history interface alone.

We use these plots to separate two kinds of failure. The first is \emph{interface failure}: the model or agent could in principle use more structured behavioral evidence, but the direct/profile/retrieval input does not present that evidence in the right form. The second is \emph{observability failure}: even strong structured signals cannot determine the label because the target action depends on hidden constraints, unobserved timing, or off-chain strategy. This distinction is important for future benchmark users deciding whether to improve retrieval, memory construction, profile construction, model calibration, or data instrumentation.

\section{Error Complementarity}
\label{app:error_complementarity}

\begin{figure}[H]
\centering
\begin{subfigure}[t]{0.4\textwidth}
\centering
\includegraphics[width=\linewidth]{pics/upset/belief_upset_bigfont_v6_nips.pdf}
\end{subfigure}
\begin{subfigure}[t]{0.4\textwidth}
\centering
\includegraphics[width=\linewidth]{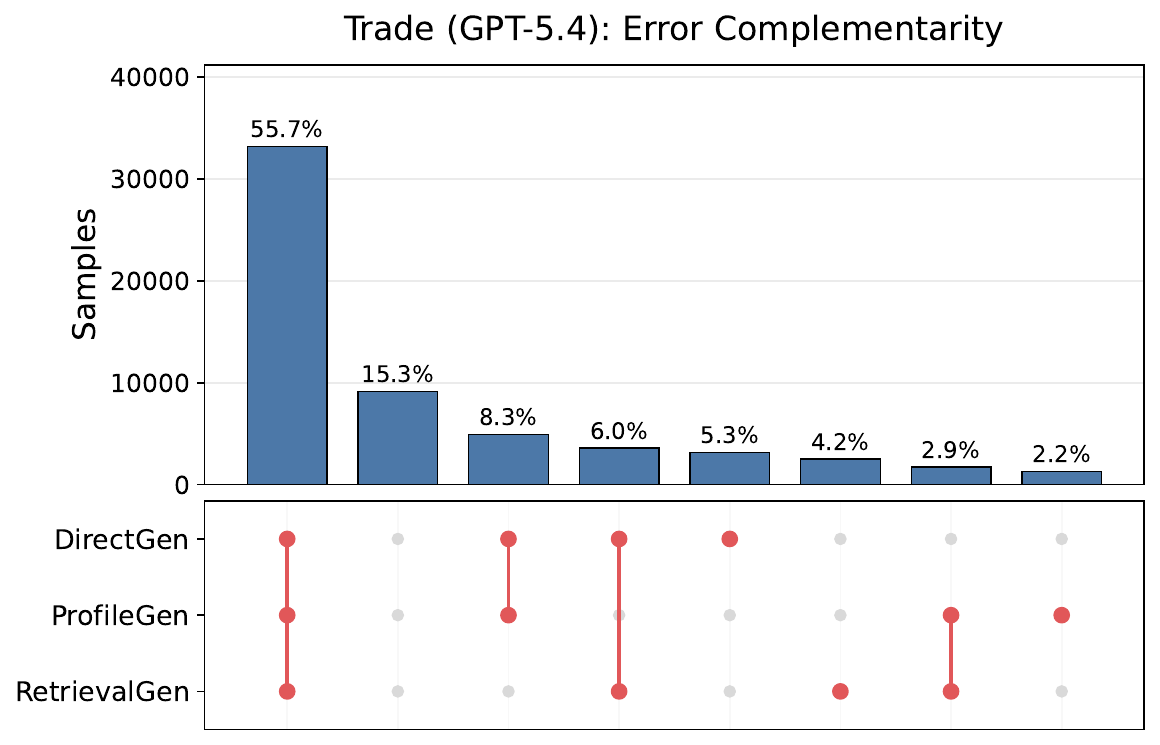}
\end{subfigure}

\caption{
Upset experiments across Belief and Trade layers.
}
\label{fig:upset_all}
\end{figure}

Figure~\ref{fig:upset_all} shows overlap in GPT-5.4 errors across DirectGen, ProfileGen, and RetrievalGen. Large shared-error regions indicate hard instances that are not solved by simply changing the history interface.

The upset analysis asks a different question from the main leaderboard: do methods fail on the same examples? If two methods have similar aggregate accuracy but make different mistakes, then they may contain complementary information. If most errors are shared, then the benchmark contains a hard subset that is robust to the tested history interfaces. This diagnostic is useful because personalization methods can appear close in average performance while behaving very differently at the instance level.

For both Belief and Trade, the shared-error mass is substantial. This means that many mistakes are not merely artifacts of choosing DirectGen, ProfileGen, or RetrievalGen. At the same time, method-specific error regions remain nontrivial, which suggests that future systems may benefit from adaptive history selection or ensembles that choose between raw recent history, profile summaries, and retrieved support evidence on a per-instance basis.

\section{History Scaling}
\label{app:history_scaling}

\begin{figure}[H]
\centering
\begin{subfigure}[t]{0.24\textwidth}
\centering
\includegraphics[width=\linewidth]{pics/history_scaling/belief_direction_acc_nips_clean_v3.pdf}
\end{subfigure}
\begin{subfigure}[t]{0.24\textwidth}
\centering
\includegraphics[width=\linewidth]{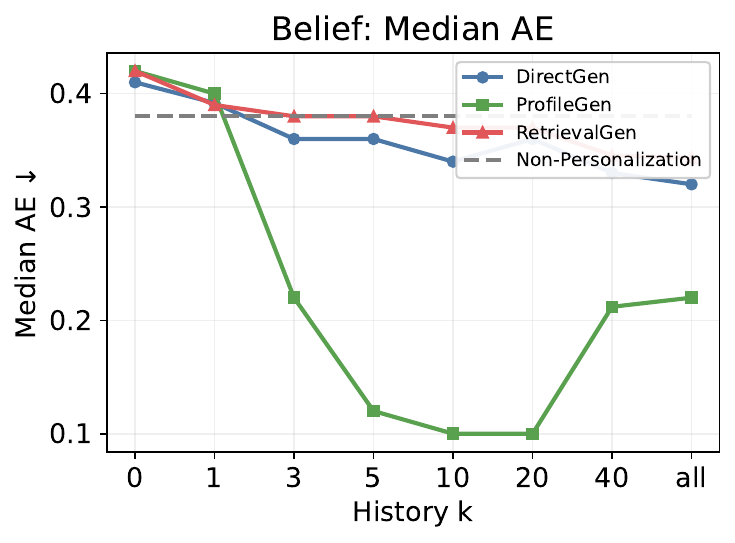}
\end{subfigure}
\begin{subfigure}[t]{0.24\textwidth}
\centering
\includegraphics[width=\linewidth]{pics/history_scaling/trade_direction_acc_nips_clean_v3.pdf}
\end{subfigure}
\begin{subfigure}[t]{0.24\textwidth}
\centering
\includegraphics[width=\linewidth]{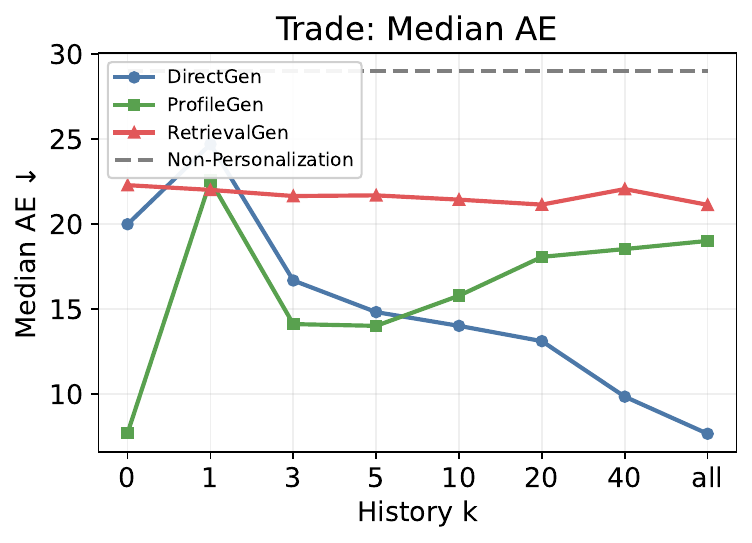}
\end{subfigure}
\caption{
History scaling experiments across both layers and metrics.
}
\label{fig:history_scaling_all}
\end{figure}

Figure~\ref{fig:history_scaling_all} reports both accuracy and absolute-error history-scaling diagnostics. The main paper discusses the direction/choice accuracy plots; the additional error plots show how calibration and amount prediction respond to increasing history size.

The history-scaling study varies how much prior behavior is made visible to the model while holding the target task fixed. This analysis is diagnostic rather than a replacement for the full-test benchmark: it is run on a 100-wallet subset so that multiple history lengths can be evaluated at manageable cost. The goal is to measure whether additional history improves prediction monotonically, saturates, or eventually hurts due to long-context drift and irrelevant evidence.

The Belief curves show that a modest amount of history can be enough to recover much of the personalization benefit. ProfileGen improves sharply when it has enough history to summarize stable preferences, but very long histories can degrade error metrics, likely because the profile must compress heterogeneous behavior into a small schema. This supports the design choice of using recency-truncated histories in the main input pipeline.

The Trade curves tell a different story. DirectGen benefits more steadily from additional history, especially for direction prediction and amount error. This suggests that local transaction prediction often needs concrete sequential evidence rather than only a high-level user summary. RetrievalGen is comparatively flat in the scaling study, implying that the support-wallet evidence used here is less effective than target-wallet sequential history for predicting individual trades.

\FloatBarrier

\end{document}